\documentclass{article}

 \usepackage[preprint]{neurips_2026}

\usepackage[utf8]{inputenc} 
\usepackage[T1]{fontenc}    
\usepackage{hyperref}       
\usepackage{url}            
\usepackage{booktabs}       
\usepackage{amsfonts}       
\usepackage{nicefrac}       
\usepackage{microtype}      
\usepackage{xcolor}         
\usepackage{amsmath}
\usepackage{comment}
\usepackage{graphicx}
\usepackage{booktabs}
\usepackage[table,dvipsnames]{xcolor}
\usepackage{array}
\usepackage{placeins}
\usepackage{float}
\usepackage{multirow}
\usepackage{float}
\usepackage{wrapfig}
\usepackage{tcolorbox}

\newcolumntype{L}[1]{>{\raggedright\arraybackslash}p{#1}}
\newcommand{\gain}[1]{\textcolor{ForestGreen}{\tiny(+#1)}}
\newcommand{\nogain}[1]{\textcolor{gray}{\tiny(#1)}}
\definecolor{svbestblue}{RGB}{226,235,249}
\newcommand{\svbest}[1]{\cellcolor{svbestblue}{#1}}
\title{MINER: Mining Multimodal Internal Representation for Efficient Retrieval}

\author{
Weien Li\textsuperscript{1, 2}\thanks{Equal contribution with random order.}\,\,\thanks{Corresponding to \href{mailto:weien.li@mail.mcgill.ca}{weien.li@mail.mcgill.ca}, \href{mailto:rui.song@mail.mcgill.ca}{rui.song@mail.mcgill.ca} or \href{mailto:ye.yuan3@mail.mcgill.ca}{ye.yuan3@mail.mcgill.ca}.}\And
Rui Song\textsuperscript{1}\footnotemark[1]\,\,\footnotemark[2]\And
Zeyu Li\textsuperscript{1}\And 
Haochen Liu\textsuperscript{3}\And 
Gonghao Zhang\textsuperscript{4}\And
Difan Jiao\textsuperscript{5}\And
Zhenwei Tang\textsuperscript{5}\And
Bowei He\textsuperscript{6}\And
Haolun Wu\textsuperscript{1, 7}\And
Xue Liu\textsuperscript{6, 1, 7}\And
Ye Yuan\textsuperscript{1, 7}\footnotemark[2]\And
\\
\textsuperscript{1} McGill University, 
\textsuperscript{2} MIT - Massachusetts Institute of Technology,\\
\textsuperscript{3} University of Cambridge,
\textsuperscript{4} flab.ai,
\textsuperscript{5} University of Toronto,\\
\textsuperscript{6} MBZUAI - Mohamed bin Zayed University of Artificial Intelligence,
\textsuperscript{7} Mila - Quebec AI Institute
}

\begin{document}

\maketitle

\begin{abstract}
Visual document retrieval has become essential for accessing information in visually rich documents.
Existing approaches fall into two camps.
Late-interaction retrievers achieve strong quality through fine-grained token-level matching but store hundreds of vectors per page, incurring large index footprints and high serving costs.
By contrast, dense single-vector retrievers retain storage and latency advantages but consistently lag in quality because they compress all information into a single final-layer embedding.
In this work, we first conduct a layerwise diagnostic on single-vector retrievers, revealing that retrieval-relevant signal resides in internal representations.
Motivated by these findings, we propose \textbf{MINER} (\textbf{M}ining Multimodal \textbf{I}nternal Represe\textbf{N}tation for \textbf{E}fficient \textbf{R}etrieval), a lightweight plug-in module that probes and fuses internal signals across transformer layers into a single compact embedding without modifying the backbone or sacrificing single-vector efficiency.
The first \emph{Retrieval-Aligned Layer Probing} stage attaches a lightweight probe at each layer, surfacing which dimensions carry retrieval-relevant information.
The subsequent \emph{Adaptive Sparse Multi-Layer Fusion} stage applies performance-adaptive neuron-level masking to the selected layers and fuses the surviving signals into the final dense vector.
Across ViDoRe V$1$/V$2$/V$3$, MINER outperforms existing dense single-vector retrievers on the majority of benchmarks, with up to 4.5\% nDCG@$5$ improvement over its corresponding backbone.
Compared to strong late-interaction baselines, in some settings MINER substantially narrows the nDCG@$5$ gap to $0.2$ while preserving the storage and serving advantages of dense retrieval.
%
\end{abstract}
\section{Introduction} \label{sec:intro}
Visually rich documents such as presentation slides, scanned reports, and scientific posters convey information through a tightly coupled combination of text, layout, and figures~\citep{ding2026deep}. 
Accessing information in such documents requires retrieval systems that can jointly reason over visual and textual cues~\citep{gao2025scaling, ma2024unifying}. 
This need has driven growing interest in visual document retrieval, where the goal is to retrieve relevant document pages directly from 
their rendered images, bypassing the fragile OCR-and-chunking pipelines that traditional text-based retrieval systems depend on~\citep{faysse2024colpali, ICLR2025_3640a199}.

Existing visual document retrieval methods largely fall into two families. 
Late-interaction retrievers, originating from ColBERT~\citep{khattab2020colbert} and extended to the multimodal setting by ColPali~\citep{faysse2024colpali}, represent each query and document as a set of token-level embeddings and compute relevance through fine-grained matching. 
This preserves rich local signals and yields strong retrieval quality, but at a substantial cost: each document page may require storing over a thousand patch-level vectors, leading to large index footprints and expensive scoring at retrieval time~\citep{ma2025storageefficientvdr, santhanam-etal-2022-colbertv2}. 
Dense single-vector retrievers take the opposite approach, compressing each input into a single embedding and scoring by a simple dot product~\citep{karpukhin2020dpr,lin2024mmembed,gunther2025jinaembeddingsv4}. 
This yields compact indices and efficient serving, but the compression into one final-layer vector creates an information bottleneck: retrieval-relevant signals presented in intermediate representations may be discarded before reaching the output embedding.

A growing body of recent work suggests that the internal representations of deep networks carry richer task-relevant information than their final outputs alone~\citep{evci2022head2toe, tu2023visual, zhang2024parameter}. 
In natural language processing, SPIN~\citep{jiao2023spin} and SIREN~\citep{ye2026siren} have demonstrated that intermediate layers encode semantically useful information that benefits downstream tasks such as text classification. 
For vision tasks, Perception Encoder~\citep{bolya2025perceptionencoder} shows that the strongest visual embeddings often emerge before the last layer rather than at the network output. 
These findings raise a natural question for multimodal retrieval: 
\begin{tcolorbox}[colback=gray!5!white, colframe=gray!30, boxrule=0.3pt, arc=2pt, left=4pt, right=4pt, top=2pt, bottom=2pt]
\textbf{\textit{Can internal representations of single-vector retrievers be leveraged to recover the retrieval-relevant information lost to the final layer bottleneck, thereby improving retrieval quality without sacrificing single-vector efficiency?}}
\end{tcolorbox}

To answer this question, we first conduct a systematic layerwise analysis of single-vector retrievers. 
Using normalized Centered Kernel Alignment (CKA)~\citep{kornblith2019cka} and a novel alignment-ratio diagnostic, we reveal that retrieval-relevant signal does reside in internal representations, but is distributed unevenly across depths.
Earlier layers contain useful but misaligned structure, while later layers are progressively more aligned with the final retrieval space. 
This analysis yields a principled criterion for selecting and partitioning the layers that are most amenable to lightweight extraction.

Motivated by these findings, we propose \textbf{MINER} (\textbf{M}ining Multimodal \textbf{I}nternal Represe\textbf{N}tation for \textbf{E}fficient \textbf{R}etrieval), a lightweight plug-in module that enhances dense retrieval embeddings by selectively extracting and fusing retrieval-relevant signals from internal layers. 
MINER operates in two stages. 
The first stage, \emph{Retrieval-Aligned Layer Probing}, attaches a lightweight shared probe at each selected layer and trains it to align per-layer text and vision readouts to their cross-modal final-layer anchors. 
Guided by the principled criterion we derived from our layerwise analysis, we select layers that contain richer retrieval-relevant signals and partition them into two sets.
Layers in the structurally directly aligned regime are probed with simple element-wise reweighting, which is referred to as \emph{Base Probing}.
In contrast, layers in the structurally less aligned regime receive a more expressive row-normalized linear projection, termed \emph{Normalized Projection Probing}. 
The second stage, \emph{Adaptive Sparse Multi-Layer Fusion}, applies performance-adaptive neuron-level masking to the probed layers and aggregates the surviving signals into a single dense embedding through a learned cross-layer weighted sum with a global bias. 
Importantly, without modifying the backbone model, MINER is agnostic to the readout mechanism (EOS-token or pooling) and preserves the storage and serving efficiency of single-vector retrievers.
Across ViDoRe V$1$/V$2$/V$3$~\citep{faysse2024colpali,mace2025vidorev2,loison2026vidorev3}, MINER consistently outperforms existing dense single-vector retrievers and, in several settings, substantially narrows the gap 
to strong late-interaction baselines.

In summary, our contributions are as follows:
\begin{enumerate}
    \item We reveal that retrieval-relevant signal in single-vector retrievers is distributed unevenly across transformer layers, and that different layers require qualitatively different extraction strategies. We introduce normalized CKA and alignment ratio as principled diagnostic tools.

    \item We propose MINER, a plug-in module guided by the layerwise diagnostic, comprising Retrieval-Aligned Layer Probing and Adaptive Sparse Multi-Layer Fusion, which improves retrieval quality without modifying the backbone or increasing storage and serving cost.

    \item MINER consistently improves dense retrieval quality across multiple backbones and the suite of ViDoRe benchmarks, achieving up to $4.5$\% nDCG@$5$ gain over the dense baseline, while matching its storage footprint and incurring only $1.07\times$ query latency overhead, substantially narrowing the gap to late-interaction methods at $42.4\times$ lower index cost.
\end{enumerate}
\section{Related Work}

We review two lines of work most relevant to MINER: visual document retrieval methods and studies on internal representations for retrieval.

Visual document retrieval has increasingly moved away from OCR-then-chunking pipelines toward direct page-image retrieval. 
There are two main camps for existing methods.
\textit{(1)} Late-interaction retrievers, originating from ColBERT~\citep{khattab2020colbert}, have been especially influential in this transition, with ColPali~\citep{faysse2024colpali} demonstrating strong retrieval performance on visually rich documents using VLM-based multi-vector representations. 
This line has been further developed through more realistic and challenging benchmarks such as ViDoRe V$2$ and ViDoRe V$3$~\citep{mace2025vidorev2,loison2026vidorev3,dong2025mmdocir,chen2025visrbench}, as well as follow-up studies on reproducibility and storage efficiency~\citep{qiao2025reproducibility,ma2025storageefficientvdr}.
In parallel, \textit{(2)} single-vector retrievers~\citep{karpukhin2020dpr} remains attractive because of its compact indexing and efficient serving. 
FLMR~\citep{lin2023flmr} explores fine-grained multimodal retrieval for retrieval-augmented visual question answering (VQA), and MM-Embed~\citep{lin2024mmembed} studies universal multimodal retrieval with multimodal LLMs. 
On the model side, Jina-embeddings-v$4$~\citep{gunther2025jinaembeddingsv4} and Eager-Embed-v$1$~\citep{EagerEmbed} represent recent dense multimodal embedding systems designed for practical retrieval. 
Compared with late interaction, these single-vector methods preserve the efficiency advantages of indexing and serving, but they typically lag in retrieval quality.

Recent works suggest that useful signals are often distributed across intermediate layers rather than concentrated only in the final output. 
Perception Encoder~\citep{bolya2025perceptionencoder} shows that strong visual embeddings may emerge before the final network output, while recent dense retrieval work demonstrates that multi-layer representations can improve retrieval quality beyond standard last-layer pooling~\citep{xie2025multilayerdpr}. 
SPIN~\citep{jiao2023spin} and SIREN~\citep{ye2026siren} further show that probing techniques for sparse selection and integration of internal neurons can improve downstream task performance.

Overall, prior works have improved multimodal retrieval either by increasing matching granularity through late interaction or by training stronger dense embedding models. 
In contrast, guided by recent studies suggesting that valuable task information resides in internal representations, our work extracts retrieval-relevant internal signals to improve compact single-vector multimodal retrieval.
\section{Layerwise Analysis of Retrieval-Relevant Internal Representations}
\label{sec:layerwise_analysis}

We now turn to the question raised in Section~\ref{sec:intro}.
To investigate this, we conduct a layerwise analysis on three single-vector retrievers: Jina-embeddings-v$4$ (Jina)~\citep{gunther2025jinaembeddingsv4}, Eager-Embed-v$1$ (Eager)~\citep{EagerEmbed} and MoCa-$3$B(MoCa)~\citep{chen2025moca}.

Formally, we consider a single-vector retriever with a backbone of $L$ transformer layers.
Given an input, let $\mathbf{H}^{(l)}$ denote its hidden states at layer $l$, and let $r(\cdot)$ be the model-specific readout operator (e.g., EOS-token extraction or pooling).
The corresponding layerwise readout is $\mathbf{x}^{(l)} = r(\mathbf{H}^{(l)}) \in \mathbb{R}^D$, where $D$ is the embedding dimension.
For each input indexed by $i$ in a paired text-vision dataset of size $N$, we denote its layer-$l$ readout as $\mathbf{x}_i^{(l)} \in \mathbb{R}^D$ and its corresponding cross-modal final-layer anchor as $\mathbf{a}_i \in \mathbb{R}^D$, defined as the paired input's final-layer readout (i.e., the text final-layer readout when input is a vision input, and vice versa).
Our analysis compares each layer-$l$ readout against this cross-modal anchor.
Two questions naturally arise from this comparison: \textit{(1)} which layers actually contain retrieval-relevant signals that are amenable to lightweight extraction? and \textit{(2)} should all such layers be treated in the same way, or do different layers call for different extraction strategies? 
We address these two questions in turn below and conclude with a principled criterion that directly motivates the design of MINER in Section~\ref{sec:method}.

\paragraph{Which layers are the most amenable to lightweight extraction?}
A natural diagnostic is to measure how structurally similar the per-layer representation space is to the final cross-modal anchor space. 
We adopt linear Centered Kernel Alignment (CKA)~\citep{kornblith2019cka}, which compares two representation spaces at the dataset level rather than pointwise, and is therefore robust to coordinate-level mismatches that are common across layers of deep networks. 
Concretely, for each layer $l$, we stack the readouts $\{\mathbf{x}_i^{(l)}\}_{i=1}^N$ into a matrix $\mathbf{X}^{(l)} \in \mathbb{R}^{N \times D}$, and likewise stack the corresponding cross-modal anchors $\{\mathbf{a}_i\}_{i=1}^N$ into $\mathbf{A} \in \mathbb{R}^{N \times D}$.
The linear CKA between the layer-$l$ representation space and the anchor space, denoted $\mathrm{CKA}_l$, is then defined as
\begin{equation}
\mathrm{CKA}_l
=
\frac{\left\| (\mathbf{X}^{(l)})^\top \mathbf{A} \right\|_F^2}
{\left\| (\mathbf{X}^{(l)})^\top \mathbf{X}^{(l)} \right\|_F
\left\| \mathbf{A}^\top \mathbf{A} \right\|_F },
\end{equation}
where $\|\cdot\|_F$ denotes the Frobenius norm. 
A high $\mathrm{CKA}_l$ indicates that the layer-$l$ representation space is structurally similar to the final retrieval space, whereas a low $\mathrm{CKA}_l$ suggests that layer $l$ either has not yet developed sufficient retrieval-relevant structure, or expresses it in a geometry too distant from the final retrieval space to be recovered with a lightweight probe.
However, using $\mathrm{CKA}_l$ directly as a layer-selection criterion is problematic because its absolute scale can vary across backbone models. 
To obtain a cutoff that is invariant to such scale differences, we min-max normalize $\mathrm{CKA}_l$ across all $L$ layers of the backbone and retain only those layers whose normalized score $\widehat{\mathrm{CKA}}_l$ exceeds a cutoff $\tau_{\mathrm{CKA}} \in [0,1]$:
\begin{equation}
\widehat{\mathrm{CKA}}_l
=
\frac{\mathrm{CKA}_l-\mathrm{CKA}_{\min}}
{\mathrm{CKA}_{\max}-\mathrm{CKA}_{\min}},
\qquad
\mathcal{S}_{\mathrm{cand}}
=
\left\{
l \in \{1,\dots,L\} : \widehat{\mathrm{CKA}}_l \geq \tau_{\mathrm{CKA}}
\right\},
\end{equation}
where $\mathrm{CKA}_{\min}$ and $\mathrm{CKA}_{\max}$ are the minimum and maximum CKA values across all $L$ layers. 
This yields a candidate set $\mathcal{S}_{\mathrm{cand}}$ of layers whose representations are structurally close enough to the final retrieval space to be amenable to lightweight extraction.
In our study, we set $\tau_{\mathrm{CKA}} = 0.6$, which retains layers whose representations are sufficiently close to the final anchor space.
We later analyze the sensitivity of MINER to this choice in Section~\ref{subsec:sensitivity}.

\begin{figure}[t]
  \centering
  \resizebox{0.85\linewidth}{!}{
  \includegraphics[width=\linewidth]{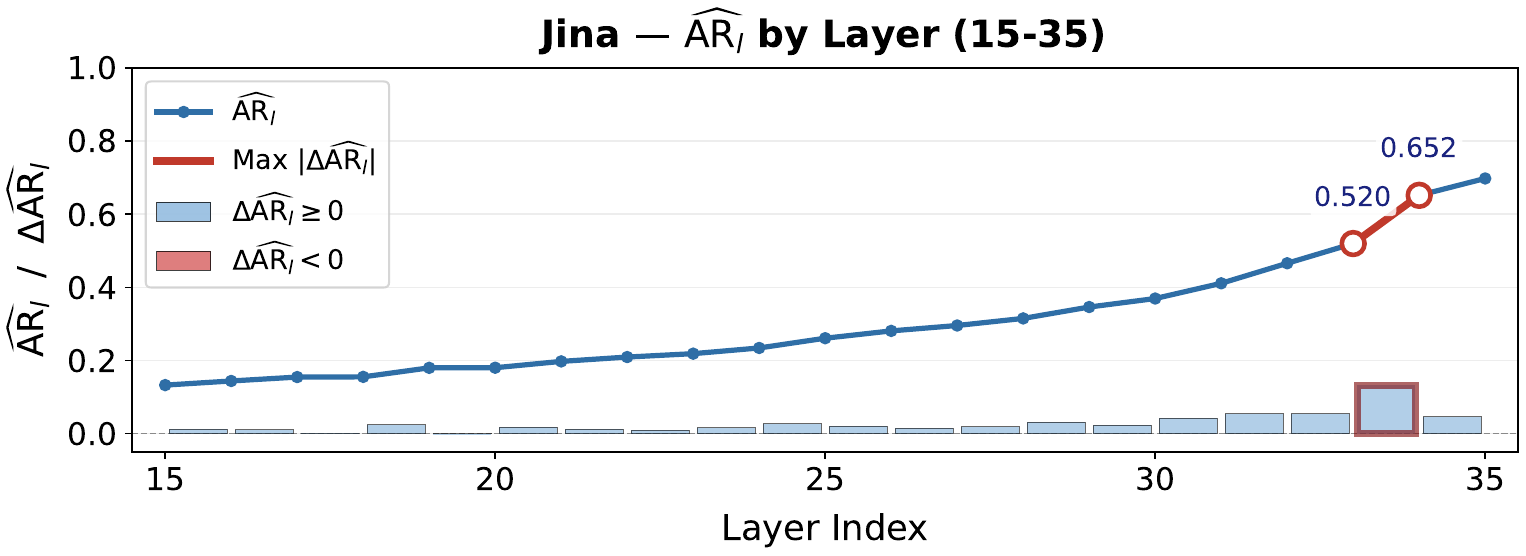}
  }
  \vspace{-2mm}
  \caption{The line shows the layerwise normalized Alignment Ratio ($\widehat{\mathrm{AR}}_l$) of Jina backbone from layer $15$ to $35$ (the candidate set $\mathcal{S}_{\mathrm{cand}}$ of layers), and the bars show $\Delta\widehat{\mathrm{AR}}_l$, the change in $\widehat{\mathrm{AR}}_l$ from layer $l-1$ to layer $l$. The largest step change occurs between layer $33$ and $34$ ($\Delta\widehat{\mathrm{AR}}_{34} = 0.1319$).}
  \label{fig:alignment_ratio_jina}
\end{figure}

\paragraph{Should all candidate layers be treated the same way?}
Structural similarity alone does not determine how easy it is to extract useful signal: a layer may share dataset-level geometry with the anchor space without being directly aligned with it at the sample level. 
This gap arises because CKA is invariant to orthogonal transformations of the representation space, while sample-level alignment is not.
To capture this distinction, we complement the dataset-level CKA with a sample-level alignment measure. 
Concretely, for each layer $l$, we compute the average per-sample cosine similarity between its readouts $\{\mathbf{x}_i^{(l)}\}_{i=1}^N$ and their corresponding cross-modal anchors $\{\mathbf{a}_i\}_{i=1}^N$, denoted as $c_l \in [-1, 1]$. 
Here, $c_l$ quantifies how well, on average, individual layer-$l$ readouts already point in the same direction as their cross-modal anchors. 
We then define the \emph{alignment ratio} of layer $l$ as $\mathrm{AR}_{l} = c_l / \mathrm{CKA}_l$. 
Intuitively, $\mathrm{AR}_l$ measures how much of a layer's structural similarity to the anchor space is already realized as direct pointwise alignment. 
A high $\mathrm{AR}_l$ indicates that the layer is not only structurally similar to the final retrieval space but also coordinate-aligned with it, so its signal can be extracted with a simple, geometry-preserving operation. 
A low $\mathrm{AR}_l$ indicates that the useful structure is present but expressed in a less aligned subspace, which calls for a more expressive transformation to realign it before fusion.
Figure~\ref{fig:alignment_ratio_jina} visualizes the min-max normalized $\widehat{\mathrm{AR}_l}$ and its adjacent-layer change $\Delta\widehat{\mathrm{AR}_l} = \widehat{\mathrm{AR}_l} - \widehat{\mathrm{AR}_{l-1}}$ for Jina, with additional visualizations for Eager and MoCa provided in Appendix~\ref{app:additional_layerwise_analysis}.
Within the candidate set $\mathcal{S}_{\mathrm{cand}}$, two distinct regimes emerge.
Earlier retained layers exhibit relatively low and slowly varying $\widehat{\mathrm{AR}_l}$, while a sharp jump in $\widehat{\mathrm{AR}_l}$ emerges starting from the last three layers, clearly visible as a large positive $\Delta\widehat{\mathrm{AR}_l}$ in Figure~\ref{fig:alignment_ratio_jina}. 
This pattern is consistent across all three backbones, as shown for Jina in Figure~\ref{fig:alignment_ratio_jina} and further supported by the Eager and MoCa visualizations in Figures~\ref{fig:alignment_ratio_eager} and~\ref{fig:alignment_ratio_moca}.
It confirms that candidate layers fall into two qualitatively different alignment regimes, motivating differentiated extraction strategies.

Together, these two diagnostics yield a principled criterion: normalized CKA selects which layers to extract from, and the alignment ratio determines how each selected layer should be treated.

\section{Mining Multimodal Internal Representations for Efficient Retrieval}
\label{sec:method}

\begin{figure}[t]
  \centering
  \includegraphics[width=\linewidth, trim=0.1cm 1.5cm 0.1cm 1.5cm, clip]{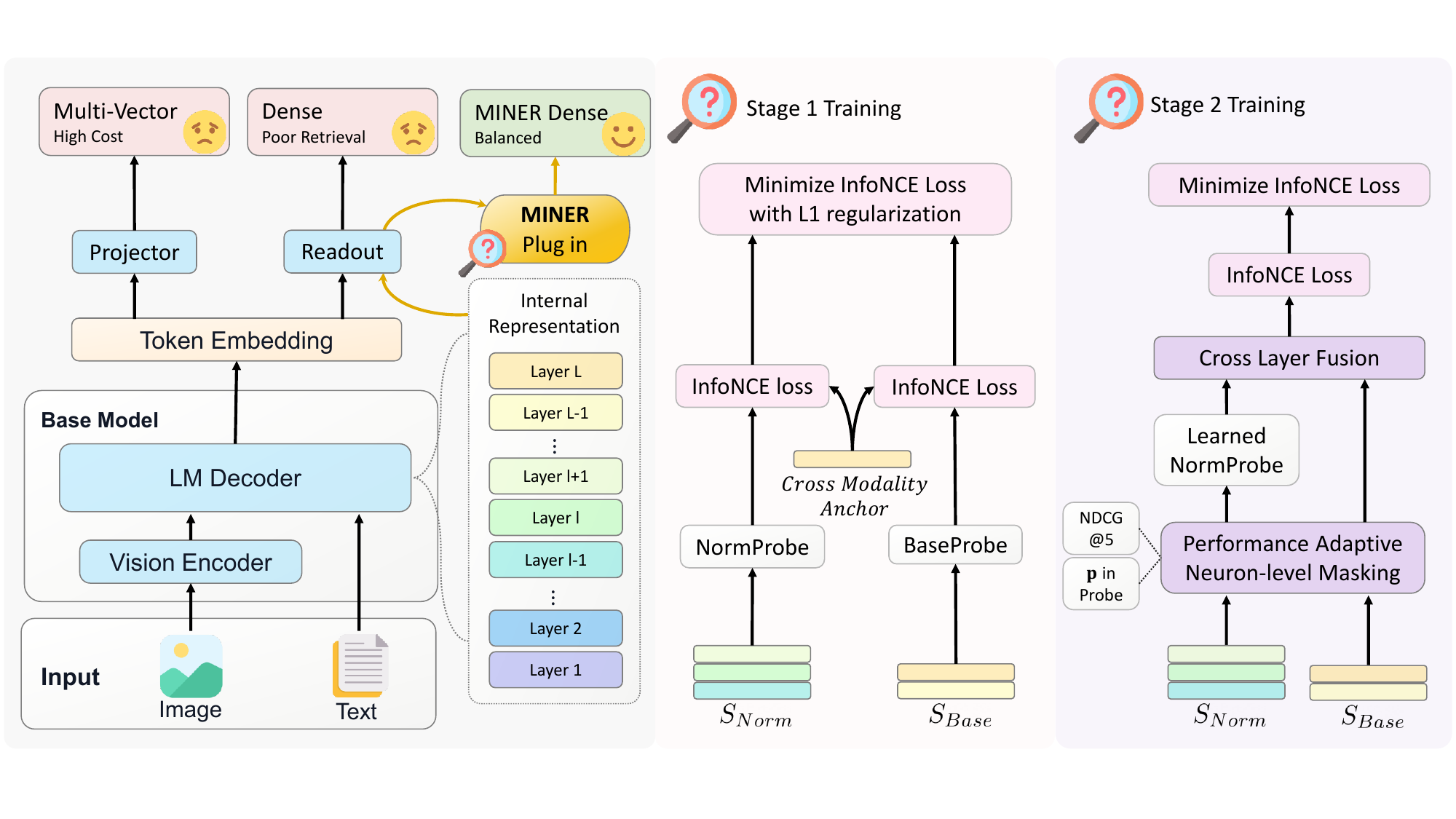}
    \vspace{-4mm}
    \caption{
    Overview of MINER. MINER acts as a plug-in module that extracts readout-aligned internal representations from the Language Model (LM) decoder and improves the final embedding.
    }
  \label{fig:miner_arch}
\end{figure}

We now formally introduce the problem we aim to address. 
In visual document retrieval, given a query $q \in \mathcal{Q}$ and a corpus of rendered document pages $\mathcal{D}$, the goal is to retrieve the most relevant document $d \in \mathcal{D}$ for $q$.
A single-vector retriever uses a backbone $f$, with $L$ transformer layers as introduced in Section~\ref{sec:layerwise_analysis}, to encode the query and document into single embeddings $\mathbf{e}_q = f(q)$ and $\mathbf{e}_d = f(d)$, and ranks documents by the inner product $s(q, d) = \mathbf{e}_q^\top \mathbf{e}_d$.
In standard single-vector retrievers, $\mathbf{e}_q$ and $\mathbf{e}_d$ are taken from the final-layer readout of $f$.
In contrast, we propose \textbf{MINER}~(\textbf{M}ining Multi-Modal \textbf{I}nternal Represe\textbf{N}tation for \textbf{E}fficient \textbf{R}etrieval), a lightweight plug-in module that constructs $\mathbf{e}_q$ and $\mathbf{e}_d$ by mining retrieval-relevant signals from multiple internal layers of $f$, guided by the criterion derived in Section~\ref{sec:layerwise_analysis}.
Figure~\ref{fig:miner_arch} illustrates the overall MINER pipeline.
MINER consists of two stages: \emph{Retrieval-Aligned Layer Probing} and \emph{Adaptive Sparse Multi-Layer Fusion}.
MINER is agnostic to the readout mechanism of $f$ (EOS-token or pooling) and does not modify $f$, thereby preserving the storage and serving advantages of single-vector retrieval.

\subsection{Retrieval-Aligned Layer Probing}
\label{subsec:probing}
Stage $1$ attaches a lightweight probe to each layer in $\mathcal{S}_{\mathrm{cand}}$ identified in Section~\ref{sec:layerwise_analysis}, with the parameterization of each probe determined by the layer's alignment regime, and trains all probes with a shared cross-modal siamese objective.
Recall from Section~\ref{sec:layerwise_analysis} the layerwise readout $\mathbf{x}^{(l)}_i \in \mathbb{R}^D$ of input $i$ at layer $l$. 
For each paired text--vision sample $(t_i, v_i)$ in our training set $\{(t_i, v_i)\}_{i=1}^N$, we use the modality-explicit notation $\mathbf{x}^{(l)}_{t_i}$ and $\mathbf{x}^{(l)}_{v_i}$ for the layer-$l$ readouts of the text and vision inputs, with the final-layer readouts $\mathbf{x}^{(L)}_{t_i}$ and $\mathbf{x}^{(L)}_{v_i}$ serving as the text and vision \emph{anchors}.
These are the cross-modal anchors $\mathbf{a}_i$ from Section~\ref{sec:layerwise_analysis}, now written in modality-explicit form.

\paragraph{Probe parameterization.}
Following the alignment-ratio analysis in Section~\ref{sec:layerwise_analysis}, we partition $\mathcal{S}_{\mathrm{cand}}$ into the last three layers, $\mathcal{S}_{\text{base}}$, which are already directly aligned with the final retrieval space, and the remaining earlier layers, $\mathcal{S}_{\text{norm}}$, which are structurally similar but expressed in a less aligned subspace. 
Each subset is equipped with its own probe parameterization, shared across the text and vision modalities since both pass through the same backbone $f$ and ultimately map into a common retrieval space.
For each layer $l \in \mathcal{S}_{\text{base}}$, we apply a \emph{Base Probe (BaseProbe)}, which learns an importance vector $\mathbf{p}_l \in \mathbb{R}^D$ and applies element-wise reweighting:
\begin{equation}
\mathrm{BaseProbe}_l(\mathbf{x}) = \mathbf{p}_l \odot \mathbf{x}.
\label{eq:base-probe}
\end{equation}
This parameterization preserves the original coordinate system and is appropriate when the layer is already coordinate-aligned with the anchor space, so only neuron-level reweighting is needed.
For each layer $l \in \mathcal{S}_{\text{norm}}$, we apply a \emph{Normalized Projection Probe (NormProbe)}, which additionally learns a projection matrix $\mathbf{W}_l \in \mathbb{R}^{D \times D}$ and applies element-wise reweighting followed by a row-normalized linear projection:
\begin{equation}
\mathrm{NormProbe}_l(\mathbf{x}) 
= \tilde{\mathbf{W}}_l (\mathbf{p}_l \odot \mathbf{x}),
\qquad
\tilde{\mathbf{W}}_l^{(j)} 
= \frac{\mathbf{W}_l^{(j)}}{\|\mathbf{W}_l^{(j)}\|_2},
\label{eq:norm-probe}
\end{equation}
where $\mathbf{W}_l^{(j)}$ denotes the $j$-th row of $\mathbf{W}_l$ and $\tilde{\mathbf{W}}_l^{(j)}$ its $\ell_2$-normalized counterpart. 
The row normalization restricts the projection to realign the less aligned subspace to the anchor space without rescaling, while $\mathbf{p}_l$ retains an explicit notion of neuron-level importance for downstream masking.

\paragraph{Siamese training objective.}
Each probe is trained to align its layer-$l$ readout with the \emph{cross-modal} final-layer anchor in a siamese manner: the probed text representation should align with the vision anchor, and vice versa.
For notational convenience, we write $\mathrm{probe}_l(\cdot)$ to denote $\mathrm{BaseProbe}_l(\cdot)$ if $l \in \mathcal{S}_{\text{base}}$ and $\mathrm{NormProbe}_l(\cdot)$ if $l \in \mathcal{S}_{\text{norm}}$.
The probe at layer $l$ then minimizes
\begin{equation}
\mathcal{L}_l =
\frac{1}{N}\sum_{i=1}^{N}
\Big[
\tfrac{1}{2}\,\ell\!\big(\mathrm{probe}_l(\mathbf{x}^{(l)}_{t_i}),\, 
\mathbf{x}^{(L)}_{v_i}\big)
+
\tfrac{1}{2}\,\ell\!\big(\mathrm{probe}_l(\mathbf{x}^{(l)}_{v_i}),\, 
\mathbf{x}^{(L)}_{t_i}\big)
\Big]
+ \lambda \|\mathbf{p}_l\|_1,
\label{eq:probe-loss}
\end{equation}
where $\ell(\cdot,\cdot)$ is the InfoNCE loss~\citep{oord2019infonce} computed over in-batch negatives and hard negatives using cosine similarity, and $\lambda \geq 0$ controls the $\ell_1$ sparsity~\citep{l1regularization} of the importance vector $\mathbf{p}_l$.

\subsection{Adaptive Sparse Multi-Layer Fusion}
\label{subsec:fusion}
After probing, MINER aggregates the per-layer signals through two steps: \emph{performance-adaptive neuron-level masking}, which retains only the most retrieval-relevant neurons within each layer, and \emph{cross-layer fusion}, which integrates the retained signals into the final single embedding.

\paragraph{Performance-adaptive neuron-level masking.}
\label{Method:Neuron Masking}
Layers in $\mathcal{S}_{\mathrm{cand}}$ differ in how much they contribute to retrieval, and stronger layers should be allowed to retain more neurons than weaker ones. 
To capture this, we compute a normalized layer utility score $\alpha_l$ for each layer $l \in \mathcal{S}_{\mathrm{cand}}$ based on its standalone validation nDCG@$5$,
$
\alpha_l = 
(\mathrm{nDCG@5}_l - \mathrm{nDCG@5}_{\min}) / 
(\mathrm{nDCG@5}_{\max} - \mathrm{nDCG@5}_{\min}),
$
where $\mathrm{nDCG@5}_{\min}$ and $\mathrm{nDCG@5}_{\max}$ are the minimum and maximum standalone nDCG@$5$ across all layers in $\mathcal{S}_{\mathrm{cand}}$.
We then convert $\alpha_l$ into a layer-specific Top-$P$ retention ratio $P_l \in (0, 1]$, defined as $P_l = \alpha_l (1 - \rho) + \rho$, where $\rho \in (0, 1]$ is a floor hyperparameter that guarantees a minimum retention ratio for every layer. 
Stronger layers receive larger $P_l$ and retain more neurons.
Using the importance vector $\mathbf{p}_l$ learned during probing, we rank the $D$ dimensions in descending order of $|\mathbf{p}_l|$ and construct a hard binary mask $\mathbf{m}_l \in \{0, 1\}^D$ that assigns $1$ to the top $\lceil P_l D \rceil$ dimensions and $0$ to the rest. 
This yields a sparse layerwise signal that preserves the most retrieval-relevant neurons while fully suppressing the others.

\paragraph{Cross-layer fusion.}
Combining the layer partition from Stage 1 and the masking above, the 
processed readout $\mathbf{h}^{(l)}$ of an input at layer $l$ is $\mathbf{h}^{(l)} = \mathbf{m}_l \odot \mathbf{x}^{(l)}$ for $ l \in \mathcal{S}_{\text{base}}$ or 
$\mathbf{h}^{(l)} = \tilde{\mathbf{W}}_l\!\left(\mathbf{m}_l \odot \mathbf{x}^{(l)}\right)$ for $l \in \mathcal{S}_{\text{norm}}$, where $\tilde{\mathbf{W}}_l$ are the parameters learned by NormProbe (Eq.~\ref{eq:norm-probe}). 
Layers in $\mathcal{S}_{\text{base}}$ contribute directly after masking, whereas 
layers in $\mathcal{S}_{\text{norm}}$ are first realigned into the anchor space before contributing.
The fusion head then learns a per-layer weight matrix $\mathbf{U} \in \mathbb{R}^{|\mathcal{S}_{\mathrm{cand}}| \times D}$, whose row $\mathbf{u}_l$ provides dimension-wise fusion weights for layer $l$, together with a global bias $\mathbf{b} \in \mathbb{R}^D$. 
The final dense embedding of an input is then
\begin{equation}
\mathbf{e} = 
\sum_{l \in \mathcal{S}_{\mathrm{cand}}} \mathbf{u}_l \odot \mathbf{h}^{(l)} 
+ \mathbf{b}.
\label{eq:fusion}
\end{equation}
At training time, the fusion head is optimized with the same cross-modal siamese objective as the probes.
Given paired samples 
$\{(t_i, v_i)\}_{i=1}^N$, we minimize
$
\mathcal{L}_{\mathrm{fusion}} =
\frac{1}{N}\sum_{i=1}^{N}
\ell\!\big(\mathbf{e}_{t_i},\, \mathbf{e}_{v_i}\big),
$
where $\mathbf{e}_{t_i}$ and $\mathbf{e}_{v_i}$ are the fused 
embeddings of the text and vision inputs computed via 
Eq.~\ref{eq:fusion}, and $\ell(\cdot,\cdot)$ is the InfoNCE loss as 
in Eq.~\ref{eq:probe-loss}.
At inference time, $\mathbf{e}$ instantiates to the fused embeddings of the query $q$ and document $d$, which are then ranked by the inner product as $s(q, d)$.

\section{Experiment}

We conduct comprehensive experiments to answer the following research questions: 
\textbf{\emph{RQ1:}} Does MINER improve upon its dense embedding backbone, and how competitive is it with representative late-interaction retrievers? 
\textbf{\emph{RQ2:}} Does MINER retain the storage and retrieval efficiency benefits as standard single-vector retrievers? 
\textbf{\emph{RQ3:}} Are the major design components of MINER necessary? 
\textbf{\emph{RQ4:}} Is MINER robust to key hyperparameters? 
We study more research questions in Appendix~\ref{app:additional_rq}.
The experiment setup is introduced in Section~\ref{subsec:experiment_setup} first, followed by the results and analysis to each of the research questions.

\begin{table}[t]
\centering
\caption{
NDCG@$5$ on ViDoRe V$2$.
\textbf{Bold} indicates the best score in each column and \underline{underline} indicates the second-best score.
\colorbox{svbestblue}{Blue shading} denotes the best single-vector retriever in each column.
$^*$ indicates that MINER has statistically significant improvement over its corresponding backbone.
}
\label{tab:vidore-v2-main}

\small
\setlength{\tabcolsep}{3.2pt}
\renewcommand{\arraystretch}{1.12}

\begin{tabular}{@{}L{3cm} L{1cm} L{1.6cm} L{1.6cm} L{1.85cm} L{1.6cm} L{1.6cm}@{}}
\toprule
\textbf{Model} & \textbf{Size} & \textbf{Bio} & \textbf{Econ} & \textbf{ESG-Human} & \textbf{ESG-Syn} & \textbf{Avg.} \\
\midrule

\multicolumn{7}{c}{\textit{Multi-Vector Retrievers}} \\
\midrule
ColPali        & 3B & $54.6$ & $48.6$ & $58.5$ & \underline{$54.9$} & $54.2$ \\
ColQwen2       & 2B & $56.3$ & $50.6$ & $60.4$ & $52.5$ & $55.0$ \\
ColQwen2.5     & 3B & $59.2$ & $53.3$ & $\mathbf{66.4}$ & $\mathbf{58.3}$ & $\mathbf{59.3}$ \\
Jina(LI)   & 4B & $60.9$ & $51.9$ & \underline{$65.1$} & $52.5$ & $57.6$ \\

\midrule
\multicolumn{7}{c}{\textit{Single-Vector Retrievers}} \\
\midrule
SigLip      & 877M & $27.7$ & $18.6$ & $44.5$ & $35.0$ & $31.4$ \\
VLM2Vec     & 4B   & $33.2$ & $23.3$ & $36.3$ & $27.8$ & $30.1$ \\
Jina    & 4B   & $58.5$ & $55.1$ & $54.8$ & $44.8$ & $53.3$ \\
Eager & 4B   & \underline{$63.4$} & $56.6$ & $51.7$ & $52.2$ & $56.0$ \\
MoCa     & 3B   & $59.7$ & $57.0$ & $63.5$ & \svbest{$53.1$} & $58.3$ \\

\midrule
\multicolumn{7}{c}{\textit{Ours: MINER Single-Vector Retriever}} \\
\midrule
\textbf{MINER-Jina}
& 4B
& $58.6$\gain{$0.2 \%$}
& $56.8$\gain{$3.1 \%$}$^*$
& $56.8$\gain{$3.7 \%$}
& $48.7$\gain{$8.7 \%$}$^*$
& $55.2$\gain{$3.6 \%$}$^*$ \\

\textbf{MINER-Eager}
& 4B
& \svbest{$\mathbf{64.1}$\gain{$1.1\%$}$^*$}
& \underline{$57.8$}\gain{$2.1 \%$}
& $59.1$\gain{$14.3 \%$}$^*$
& $52.9$\gain{$1.4 \%$}
& $58.5$\gain{$4.5 \%$}$^*$ \\

\textbf{MINER-MoCa}
& 3B
& $59.2$\nogain{$-0.8 \%$}
& \svbest{$\mathbf{59.6}$\gain{$4.6 \%$}$^*$}
& \svbest{$64.8$\gain{$2.1 \%$}}
& $52.9$\nogain{$-0.4 \%$}
& \svbest{\underline{$59.1$}\gain{$1.4 \%$}} \\

\bottomrule
\end{tabular}
\end{table}

\subsection{Experiment Setup} \label{subsec:experiment_setup}
We evaluate MINER on three visual document retrieval backbones: Jina-Embeddings-v4 (Jina)~\citep{gunther2025jinaembeddingsv4}, Eager-Embed-v1 (Eager)~\citep{EagerEmbed}, and MoCa-3B (MoCa)~\citep{chen2025moca}. 
Jina supports both dense retrieval and late interaction, enabling a direct comparison between dense, late-interaction, and MINER-enhanced dense embeddings under the same backbone. 
Eager supports dense single-vector retrieval.
MoCa is a stronger bidirectional multimodal embedding backbone obtained through modality-aware continual pre-training followed by heterogeneous contrastive fine-tuning. 
The three backbones differ in their readout mechanisms: Eager uses an EOS-token readout, whereas Jina and MoCa use pooling-based readouts. 
This allows us to test whether MINER generalizes across different readout designs.
For each backbone, we train the MINER plug-in module using the same training data used by the corresponding backbone.
This ensures that any performance gain comes from better utilization of the model's internal representations rather than from introducing additional data. 
As for the Performance-adaptive neuron-level masking introduced in \ref{Method:Neuron Masking} it uses the training data's corresponding validation set. 
Additional implementation details are provided in Appendix~\ref{app:Implementation_Details}.

Following GQR~\citep{uzan2026guided}, we evaluate on the ViDoRe benchmark suite, including ViDoRe V$1$~\citep{faysse2024colpali}, V$2$~\citep{mace2025vidorev2}, and V$3$~\citep{loison2026vidorev3}, which is a comprehensive visual document retrieval benchmark suite covering diverse document types, layouts, and query styles.
We follow the default evaluation protocol of each benchmark: ViDoRe V$1$ and V$2$ are evaluated using nDCG@$5$~\citep{jarvelin2002cumulated}, while ViDoRe V$3$ is evaluated using nDCG@$10$.
We use ViDoRe V$2$ as the primary benchmark because ViDoRe V$1$ is increasingly saturated in recent work, while ViDoRe V$3$ is relatively new and currently includes fewer established baselines. 
For fair comparison, we include single-vector retrievers with fewer than $5$B parameters that appear on the official ViDoRe leaderboard\footnote{\url{https://huggingface.co/spaces/vidore/vidore-leaderboard}}. 
Among multi-vector late-interaction retrievers, we report the strongest representative model from each ViDoRe-developed backbone family.
Brief descriptions of the baseline models used in our comparisons are provided in Appendix~\ref{app:Baseline_info}.

\subsection{Results}
Table~\ref{tab:vidore-v2-main} presents the results on ViDoRe V$2$, while the ViDoRe V$1$ and V$3$ result tables are provided in Appendix~\ref{app:Additional_ViDoRe} (Table~\ref{tab:vidore-v1-main} and Table~\ref{tab:vidore-v3-main}). 
Unless otherwise stated, all MINER results use the same sparsity floor $\rho = 0.2$ and CKA cutoff $\tau_{\mathrm{CKA}} = 0.6$ across backbones. 
The statistical significance is evaluated using a paired $t$-test over per-query retrieval scores between our proposed MINER and its corresponding backbone, where we use the significance level of $\alpha = 0.05$.

Across backbones and benchmark versions, MINER generally improves over its corresponding backbone. 
On ViDoRe V$2$, MINER improves Jina for a $3.6\%$ relative gain, and improves Eager for a $4.5\%$ relative gain.
MINER-MoCa also improves the average score from $58.3$ to $59.1$, a $1.4\%$ relative gain. 
Although the average improvement is not significant under the default hyperparameter setting, in our sensitivity analysis (presented later in Section~\ref{subsec:sensitivity}), varying the sparsity floor $\rho$ yields significant improvements for MoCa as well, indicating that different backbones or training strategies may shift the optimal hyperparameter regime.
The gains are also consistent on ViDoRe V$3$, where MINER improves all three backbones in average nDCG@$10$. 
It is noticeable that MINER-Jina achieves the best average single-vector score among the compared single-vector retrievers. 
On ViDoRe V$1$, even though many subsets are already saturated, MINER-Eager improves its dense backbone from $82.5$ to $84.8$ nDCG@$5$, while MINER-Jina and MINER-MoCa remain close to their strong backbones.

Overall, these results answer \textbf{\emph{RQ1}} affirmatively: MINER demonstrates its effectiveness across multiple backbones and benchmarks, and in several cases narrows the gap between single-vector retrievers and representative late-interaction retrievers.

\subsection{Efficiency Analysis}
\label{sec:efficiency}

\begin{figure}[t!]
    \centering
    \includegraphics[width=\linewidth]{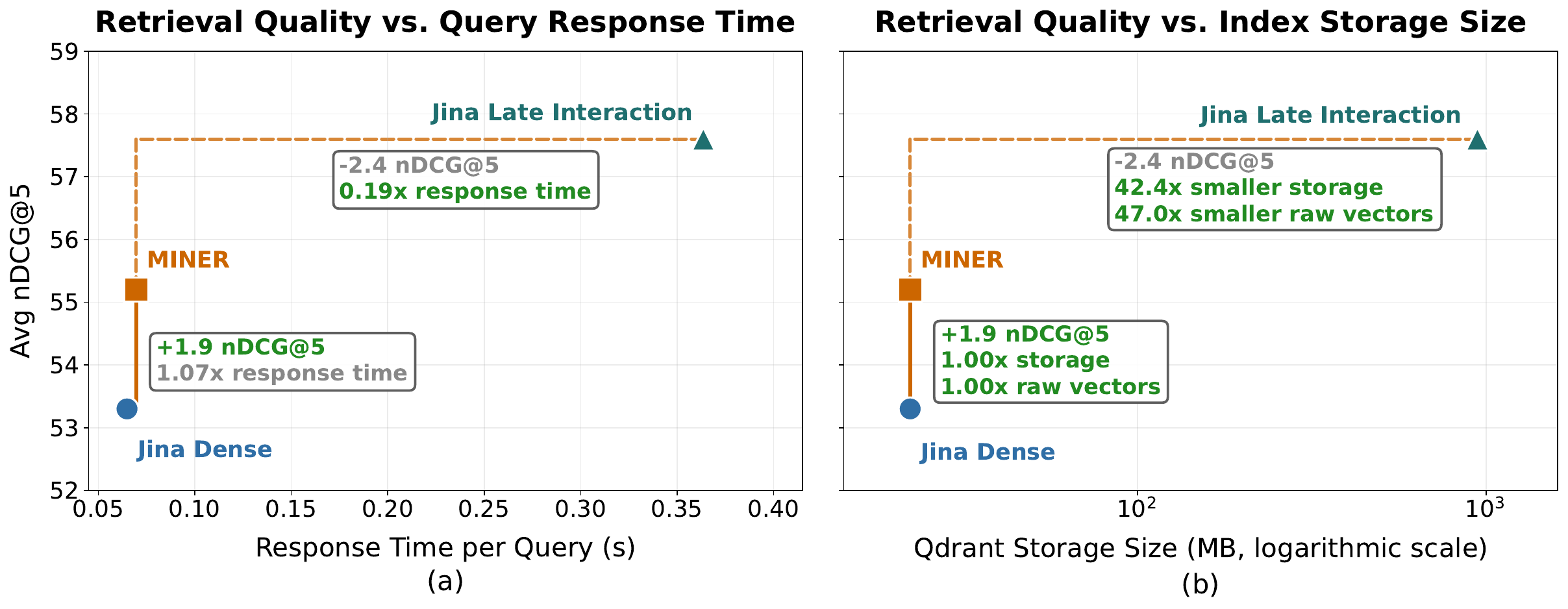}
    \vspace{-8mm}
    \caption{Efficiency-performance trade-off on ViDoRe V$2$ using Jina, by comparing dense retrieval, late interaction, and MINER across average nDCG@$5$, query response latency, and storage size.}
    \label{fig:efficiency}
\end{figure}

We evaluate the trade-off between retrieval performance, query-time latency, and storage cost. 
All efficiency measurements are conducted using Qdrant as the vector database on the full ViDoRe V$2$ benchmark. 
For each dataset, we index the complete document corpus and measure the average response time required to retrieve results for all queries. 
We report the average query response time and the average storage size. 
Full results are provided in Appendix~\ref{app:Efficiency}.

Figure~\ref{fig:efficiency} compares dense retrieval, late interaction, 
and MINER on Jina, which supports both retrieval modes under the same 
backbone. 
MINER produces a single embedding per query/document with the same dimensionality as the dense backbone, so its storage size and search complexity are identical to standard dense retrieval; the only overhead is computing the fused query embedding, resulting in a marginal end-to-end latency increase while improving retrieval quality by $+1.9$ nDCG@$5$. 
Compared to late interaction, MINER is $5.3\times$ faster and requires over $40\times$ less index storage, narrowing the quality gap with a substantially better efficiency profile. 
These results answer \textbf{\emph{RQ2}}: MINER retains the storage and retrieval efficiency of single-vector dense embeddings while recovering a large portion of the dense-to-late-interaction performance gap.

\subsection{Ablation Studies}
\label{sec:ablation}
\begin{wraptable}{r}{7.2cm}
\vspace{-10mm}
\centering
\caption{
Ablation study on ViDoRe V$2$.
}
\small
\setlength{\tabcolsep}{2.5pt}
\label{tab:ablation-v2}
\renewcommand{\arraystretch}{1.1}
\begin{tabular}{@{}lcccc@{}}
\toprule
\textbf{Backbone} & \textbf{All Neurons} & \textbf{All Base} & \textbf{All Norm} & \textbf{MINER} \\
\midrule
Jina        & 54.6 & 54.9 & 50.0 & \textbf{55.2} \\
Eager     & 57.1 & 57.4 & 53.1 & \textbf{58.5} \\
MoCa        & 58.4 & 58.4 & 56.0 & \textbf{59.1} \\
\bottomrule
\end{tabular}
\vspace{-4mm}
\end{wraptable}

To answer \textbf{\emph{RQ3}}, we ablate the major components of MINER on ViDoRe V$2$. 
\emph{All Neurons} directly trains the cross-layer weighted-sum fusion head with bias over all dimensions of the selected internal-layer readouts. 
\emph{All Base} removes the alignment-ratio partition by using BaseProbe-derived Top-$P$ masks.
\emph{All Norm} removes the alignment-ratio partition in the opposite direction by applying NormProbe to all selected layers.
These variants use the same frozen backbone, selected internal layers, and training data as the complete MINER, but remove specific components.

As shown in Table~\ref{tab:ablation-v2}, \emph{All Neurons} already improves over the corresponding single-vector baselines in Table~\ref{tab:vidore-v2-main} across all three backbones, confirming that internal layers carry useful retrieval signal.
However, this ablated variant consistently underperforms full MINER, showing that the gains are not obtained simply by fusing all dimensions.
The two probing variants further play complementary roles: \emph{All Base} remains below full MINER, while \emph{All Norm} performs worse, indicating that the normalized projection helps layers requiring additional alignment but distorts already aligned representations. 
Together, these ablations answer \textbf{\emph{RQ3}}: both sparse masking and the design of BaseProbe and NormProbe are necessary. 
We also visualize the final neuron-selection in Appendix~\ref{app:Neuron_select_plots}.

\subsection{Sensitivity Analysis}
\label{subsec:sensitivity}

\begin{figure}[t]
    \centering
    \includegraphics[width=\linewidth]{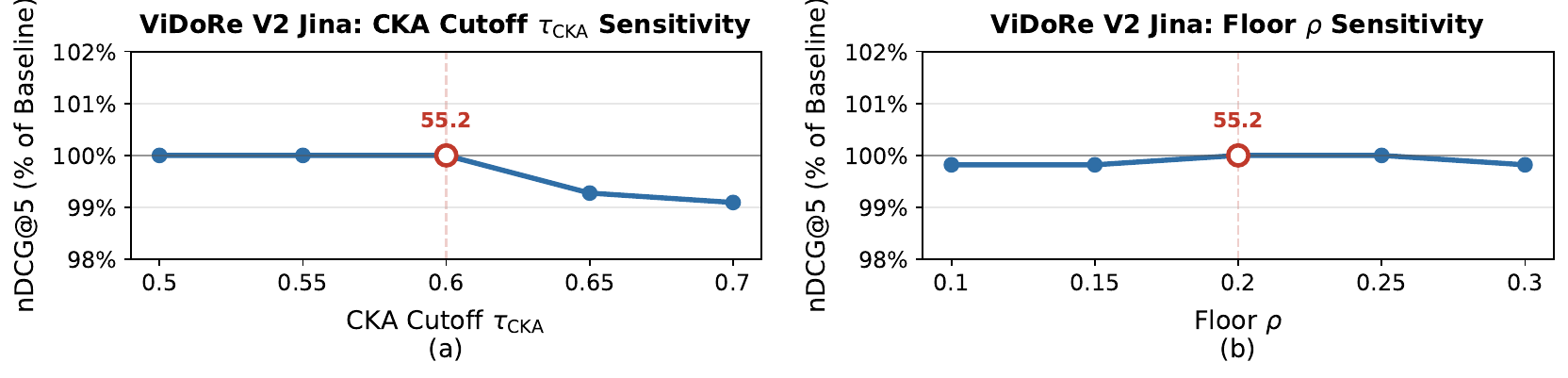}
    \vspace{-8mm}
    \caption{
    Hyperparameters sensitivity analysis, reported as a percentage of the default configuration.
    }
    \label{fig:jina_sensitivity_checks}
\end{figure}

We finally evaluate whether MINER is robust to key hyperparameters. 
We focus on two hyperparameters: the CKA cutoff $\tau_{\mathrm{CKA}}$ for selecting candidate layers and the sparsity floor $\rho$ used in dynamic Top-$P$ masking. 
We conduct the sensitivity analysis on MINER-Jina over ViDoRe V$2$ and report performance as a percentage of the default setting, where the default configuration uses CKA cutoff $0.6$ and floor $\rho=0.2$.

As shown in Figure~\ref{fig:jina_sensitivity_checks}, MINER remains stable across a wide range of CKA cutoffs $\tau_{\mathrm{CKA}}$ and sparsity floors $\rho$, indicating that the method is not highly sensitive to the exact number of selected layers or to the precise retention floor. 
We observe the same trend on MINER-Eager and MINER-MoCa.
Additional sensitivity analysis plots for MINER with these two are provided in Appendix~\ref{app:sensitivity}. 
These results answer \textbf{\emph{RQ4}} affirmatively: MINER is robust to the key hyperparameters.

\section{Conclusion and Discussion}

We presented MINER, a lightweight plug-in module for improving dense visual document retrieval by extracting retrieval-useful information from internal LM-decoder representations.
Across multiple backbones and ViDoRe benchmarks, MINER consistently improves dense retrieval quality, narrows the gap to late-interaction methods, and preserves the storage and retrieval efficiency advantages of dense embeddings.

Future work can extend MINER to stronger multimodal embedding backbones, broader retrieval domains, and settings where training data access is limited. 
Another promising direction is to develop more automatic strategies for selecting layers, probes, and sparsity levels, further reducing the need for validation-based hyperparameter choices.
We discuss the limitation in Appendix~\ref{app:limitation}.

\newpage
\bibliographystyle{unsrt}
\bibliography{references}

\newpage
\appendix

\clearpage
\section{Additional Layer-wise Analysis}
\label{app:additional_layerwise_analysis}

\begin{figure}[H]
  \centering
  \includegraphics[width=\linewidth]{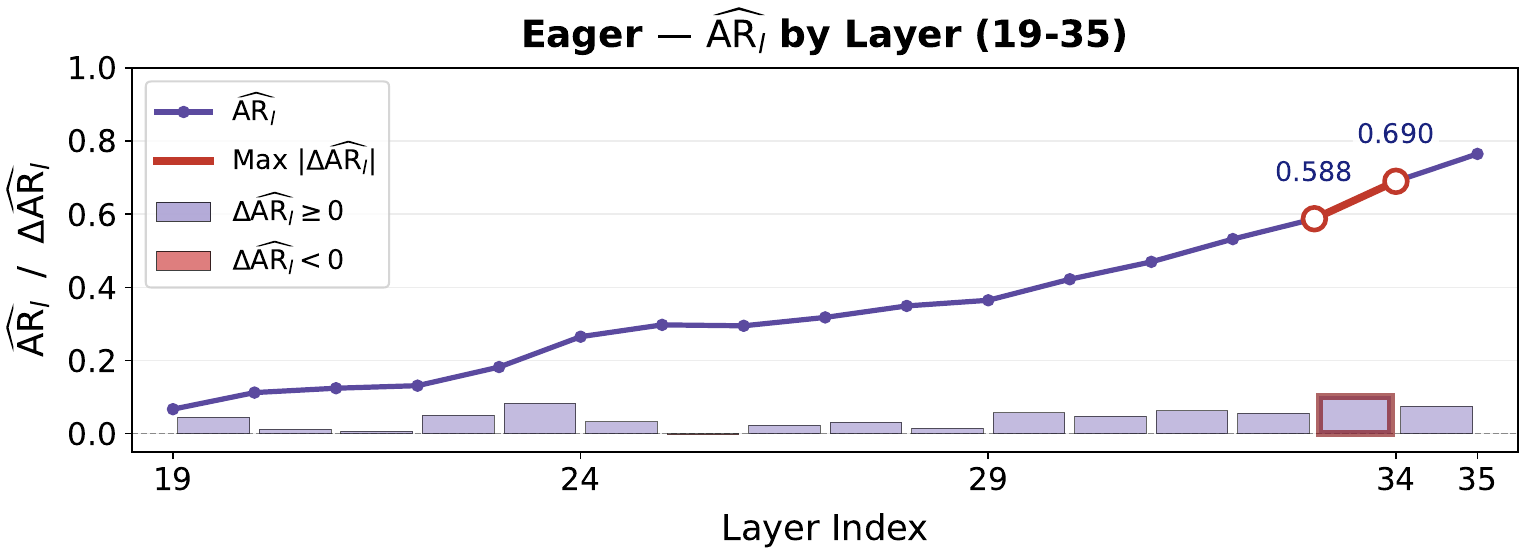}
  \caption{The line shows the layerwise normalized Alignment Ratio ($\widehat{\mathrm{AR}}_l$) of Eager backbone from layer 19 to 35 (the candidate set $S_{cand}$ of layers), and the bars show $\Delta\widehat{\mathrm{AR}}_l$, the change in $\widehat{\mathrm{AR}}_l$ from layer $l-1$ to layer l. The largest step occurs between layers 33 and 34 ($\Delta \widehat{\mathrm{AR}}_{34} = 0.1023$).}
  \label{fig:alignment_ratio_eager}
\end{figure}

\begin{figure}[H]
  \centering
  \includegraphics[width=\linewidth]{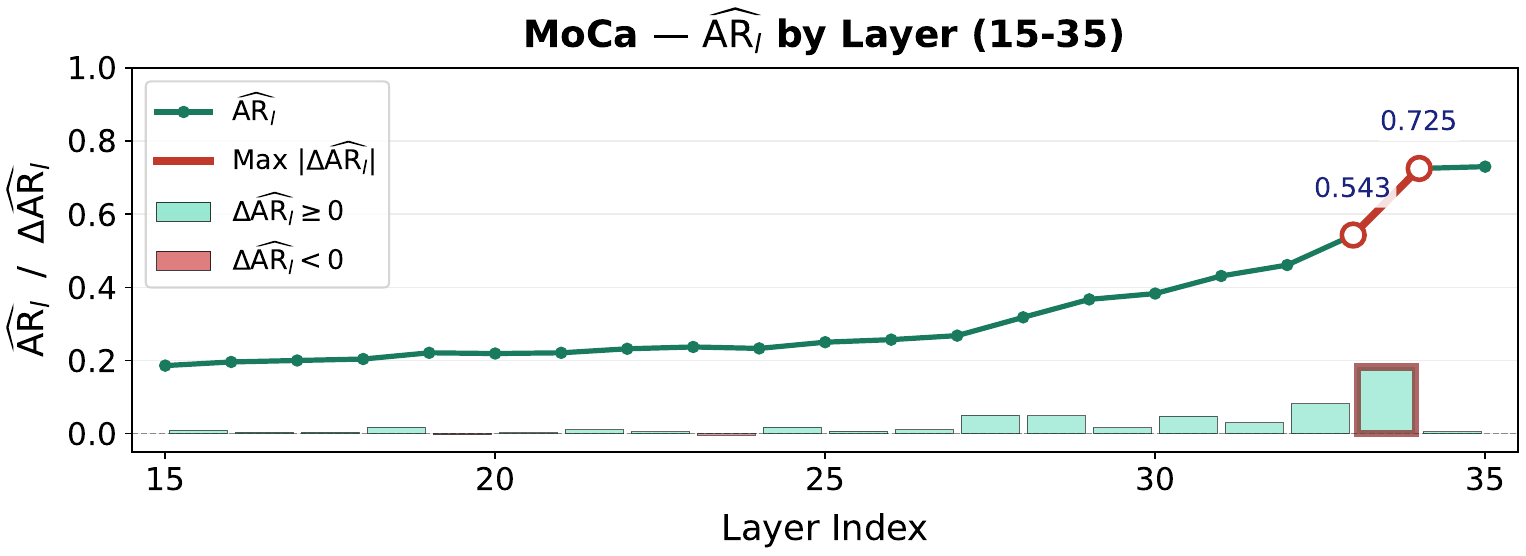}
  \caption{The line shows the layerwise normalized Alignment Ratio ($\widehat{\mathrm{AR}}_l$) of MoCa backbone from layer 15 to 35(the candidate set $S_{cand}$ of layers), and the bars show $\Delta \widehat{\mathrm{AR}}_l$, the change in $\widehat{\mathrm{AR}}_l$ from layer $l-1$ to layer l. The largest single-step change, which occurs between layers 33 and 34 ($\Delta \widehat{\mathrm{AR}}_{34} = 0.1820$)}
  \label{fig:alignment_ratio_moca}
\end{figure}

\section{Additional Research Questions} \label{app:additional_rq}

\textbf{\emph{RQ5:}} Does feature misalignment occur across internal layers, and does the proposed alignment ratio reliably characterize it?

\begin{figure}[H]
  \centering
  \resizebox{0.75\linewidth}{!}{%
    \includegraphics{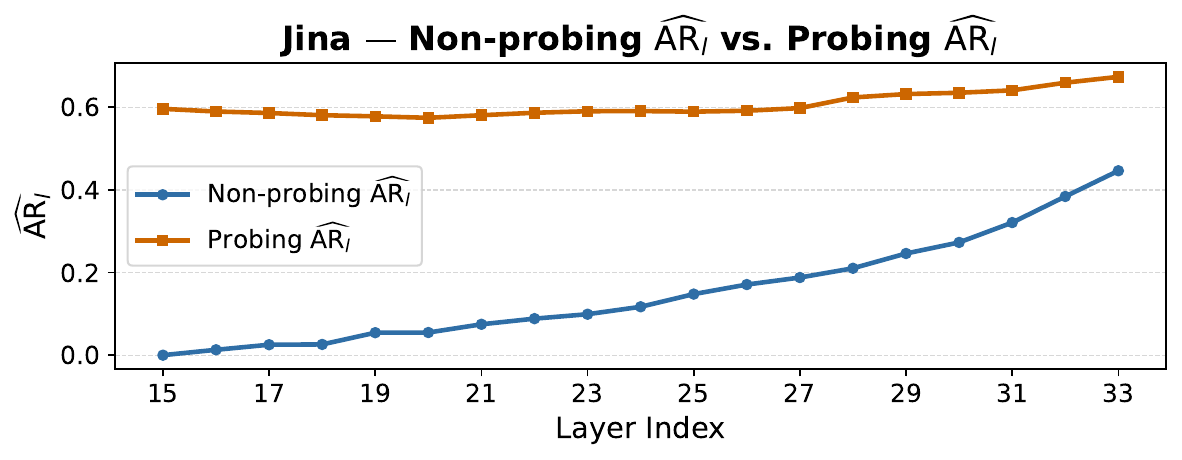}%
  }
  \vspace{-3mm}
  \caption{
  Layerwise normalized alignment ratio $\widehat{\mathrm{AR}}_l$ before and after retrieval-aligned probing on Jina. Probing consistently increases $\widehat{\mathrm{AR}}_l$ across the analyzed internal layers, indicating that intermediate representations become more aligned with the final cross-modal retrieval space after the learned alignment step.
  }
  \label{fig:jina_nonprobing_vs_probing}
\end{figure}

\begin{figure}[H]
  \centering
  \resizebox{0.75\linewidth}{!}{%
    \includegraphics{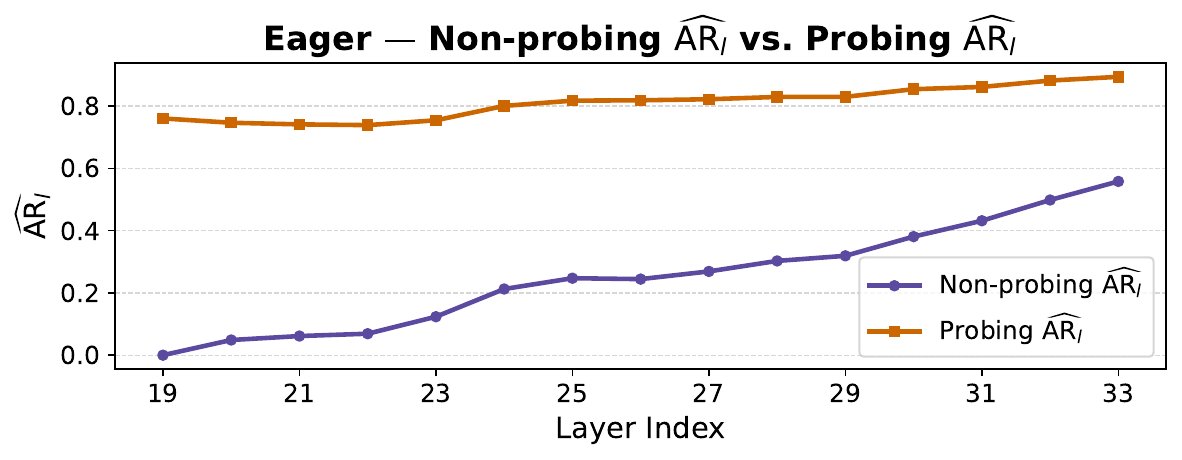}%
  }
  \caption{
  Layerwise normalized alignment ratio $\widehat{\mathrm{AR}}_l$ before and after retrieval-aligned probing on Eager. Probing consistently increases $\widehat{\mathrm{AR}}_l$ across the analyzed internal layers, indicating that intermediate representations become more aligned with the final cross-modal retrieval space after the learned alignment step.
  }
  \label{fig:eager_nonprobing_vs_probing}
\end{figure}

\begin{figure}[H]
  \centering
  \resizebox{0.75\linewidth}{!}{%
    \includegraphics{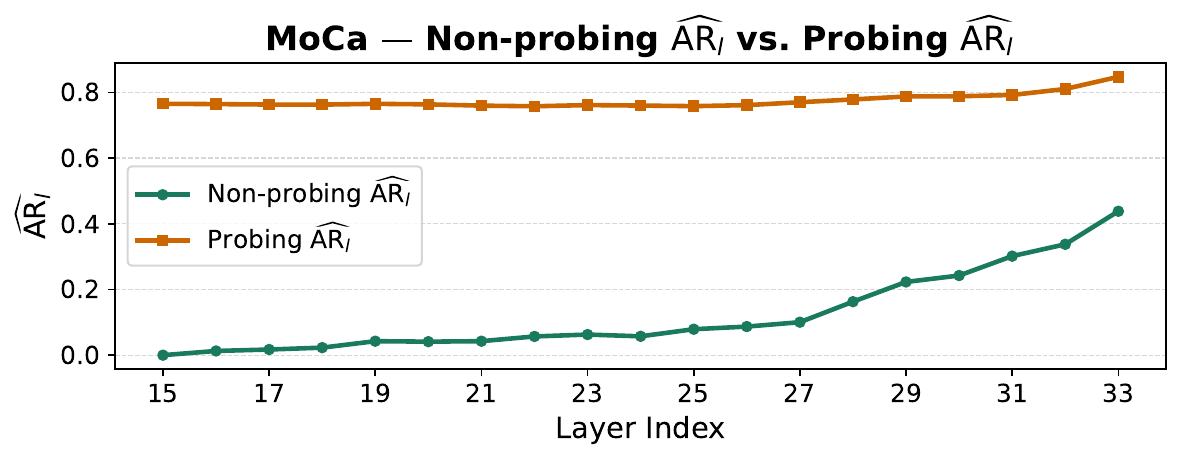}%
  }
  \caption{
  Layerwise normalized alignment ratio $\widehat{\mathrm{AR}}_l$ before and after retrieval-aligned probing on MoCa. Probing consistently increases $\widehat{\mathrm{AR}}_l$ across the analyzed internal layers, indicating that intermediate representations become more aligned with the final cross-modal retrieval space after the learned alignment step.
  }
  \label{fig:moca_nonprobing_vs_probing}
\end{figure}

We evaluate whether the proposed alignment ratio captures the expected change in layerwise feature alignment after retrieval-aligned probing.
Figure~\ref{fig:jina_nonprobing_vs_probing}, Figure \ref{fig:eager_nonprobing_vs_probing} and Figure \ref{fig:moca_nonprobing_vs_probing} compare the normalized alignment ratio $\mathrm{AR}_n$ before and after probing on their respective backbone. 
Across different backbones and among the analyzed internal layers, probing consistently increases $\mathrm{AR}_n$, indicating that the learned probes make intermediate representations more aligned with the final cross-modal retrieval space.
useful retrieval information exists in internal layers, but it may be misaligned relative to the final embedding space. 
The consistent increase in $\mathrm{AR}_n$ after probing suggests that our NormProbe can effectively re-align some of the misaligned layers to the final embedding space.

\section{Training Details}
\label{app:Implementation_Details}
\begin{table}[H]
\centering
\caption{Training details for the probing stage.}
\label{tab:probing_training_details}
\small
\setlength{\tabcolsep}{5pt}
\begin{tabular}{lcccccccc}
\toprule
\textbf{Backbone} 
& \textbf{Probes} 
& \textbf{Batch} 
& \textbf{Epochs}
& \textbf{Warm up}
& \textbf{LR} 
& $\boldsymbol{\lambda_{\ell_1}(p_\ell)}$
& \textbf{GPUs}
& \textbf{Time} \\
\midrule
Jina   & Base + WNorm & 1024 & 40 & $1$ epoch & $2 \times 10^{-4}$ & $3 \times 10^{-4}$ & $2 \times$ RTX 4090 & 17h \\
Eager  & Base + WNorm & 1024 & 40 & $1$ epoch & $2 \times 10^{-4}$ & $3 \times 10^{-4}$ & $2 \times$ RTX 4090 & 14h \\
MoCa      & Base + WNorm & 1024 & 40 & $1$ epoch & $2 \times 10^{-4}$ & $5 \times 10^{-4}$ & $2 \times$ RTX 4090 & 11h \\
\bottomrule
\end{tabular}
\end{table}

All probing models are trained with AdamW~\citep{loshchilov2019decoupled} for 40 epochs using a batch size of 1024, a learning rate of $2 \times 10^{-4}$, and random seed 42. 
We use a linear warmup for the first epoch and keep the learning rate fixed afterward. 
The weight decay is set to $0.01$ for all trainable parameters except the Hadamard scaling vector $p_\ell$, which is instead regularized by the $\ell_1$ penalty reported in Table~\ref{tab:probing_training_details}. 
Base and WNorm probes for each backbone are trained jointly with distributed data parallelism on two RTX 4090 GPUs, and the reported training cost corresponds to wall-clock time.

\begin{table}[H]
\centering
\caption{Training details for the aggregation head.}
\label{tab:head_training_details}
\small
\setlength{\tabcolsep}{5pt}
\resizebox{\textwidth}{!}{%
\begin{tabular}{lcccccccccc}
\toprule
\textbf{Backbone} 
& \textbf{CKA} 
& $\boldsymbol{S_{\text{norm}}}$ 
& $\boldsymbol{S_{\text{base}}}$ 
& \textbf{Batch} 
& \textbf{Epochs}
& \textbf{Warm up}
& \textbf{LR} 
& \textbf{Floor}
& \textbf{GPUs}
& \textbf{Time} \\
\midrule
Jina   & $0.6$ & $15$--$33$ & $34$--$36$ & $512$ & $40$ & $1$ epoch & $1 \times 10^{-5}$ & $0.2$ & $1 \times$ RTX 4090 & 8h \\
Eager  & $0.6$ & $19$--$33$ & $34$--$36$ & $512$ & $40$ & $1$ epoch & $1 \times 10^{-5}$ & $0.2$ & $1 \times$ RTX 4090 & 5h \\
MoCa      & $0.6$ & $15$--$33$ & $34$--$36$ & $512$ & $40$ & $1$ epoch & $1 \times 10^{-5}$ & $0.2$ & $1 \times$ RTX 4090 & 6h \\
\bottomrule
\end{tabular}
}
\end{table}

After AR analysis, we partition $\mathcal{S}_{\mathrm{cand}}$ into $\mathcal{S}_{\mathrm{norm}}$ and $\mathcal{S}_{\mathrm{base}}$ for all three models as shown in  Table~\ref{tab:head_training_details}.  All aggregation heads are trained with AdamW~\citep{loshchilov2019decoupled} for 40 epochs using a batch size of 512, a learning rate of $1 \times 10^{-5}$, and random seed 42. 
We use a linear warmup during the first epoch, after which the learning rate remains fixed. 
Weight decay is set to $1 \times 10^{-4}$ for all trainable parameters except the bias term $\mathbf{b}$. 
Each head is trained on a single RTX 4090 GPU, and training cost is reported as wall-clock time.
For consistency, we train all heads for the same number of epochs, despite possible differences in convergence speed.

\section{Baseline Information}
\label{app:Baseline_info}
For completeness, we briefly describe the baseline models included in our comparisons. 

\paragraph{SigLIP.}
SigLIP is a CLIP-style vision-language model trained with a sigmoid pairwise loss rather than a softmax contrastive loss, enabling image-text representation learning without global batch-level normalization; we use the shape-optimized SoViT-400M variant pretrained on WebLI at $384 \times 384$ resolution~\citep{zhai2023sigmoid}.

\paragraph{VLM2Vec.}
VLM2Vec is an instruction-guided multimodal embedding model that converts a pretrained vision-language model into a fixed-dimensional embedding model through contrastive training on the MMEB benchmark; we use the full TIGER-Lab VLM2Vec checkpoint in our dense-retrieval comparison~\citep{jiang2024vlm2vec}.

\paragraph{Jina-Embeddings-v4 (Jina).}
Jina-Embeddings-v4 is a $4$B-parameter multimodal and multilingual embedding model built on Qwen2.5-VL-3B-Instruct, supporting both dense single-vector retrieval and late-interaction multi-vector retrieval for text, images, and visually rich documents~\citep{gunther2025jinaembeddingsv4}.

\paragraph{Eager-Embed-v1 (Eager).}
Eager-Embed-v1 is a $4$B-parameter multimodal dense embedding model finetuned from Qwen3-VL-4B-Instruct for efficient visual document retrieval, using single-vector embeddings to index documents without the max-sim scoring used by ColBERT-style multi-vector retrievers~\citep{EagerEmbed}.

\paragraph{MoCa-3B (MoCa).}
MoCa-3B is a bidirectional multimodal embedding model trained from Qwen2.5-VL-3B-Instruct using modality-aware continual pre-training followed by heterogeneous contrastive fine-tuning, supporting text, image, and interleaved multimodal inputs for general multimodal retrieval and visual document retrieval~\citep{chen2025moca}.

\paragraph{ColPali.}
ColPali is a PaliGemma-3B-based visual document retriever that produces ColBERT-style multi-vector representations of queries and document images, using late interaction between text tokens and image patches for fine-grained visual document retrieval~\citep{faysse2024colpali}.

\paragraph{ColQwen2.}
ColQwen2 extends Qwen2-VL-2B-Instruct into a ColBERT-style late-interaction retriever, generating multi-vector representations for both text queries and document images and scoring them through token-level MaxSim-style matching~\citep{faysse2024colpali}.

\paragraph{ColQwen2.5.}
ColQwen2.5 is a Qwen2.5-VL-3B-Instruct-based late-interaction visual document retriever that generates ColBERT-style multi-vector representations of text queries and document images, using dynamic image resolutions with up to $768$ image patches for fine-grained document matching~\citep{faysse2024colpali}.
\section{Additional ViDoRe Results}
\label{app:Additional_ViDoRe}

\begin{table*}[!h]
\centering
\caption{
NDCG@$5$ on ViDoRe V$1$. 
\textbf{Bold} indicates the best score in each column and 
\underline{underline} indicates the second-best score. 
\colorbox{svbestblue}{Blue shading} denotes the best single-vector retriever in each column.
$^*$ indicates that MINER has statistically significant improvement over its corresponding backbone.
}
\label{tab:vidore-v1-main}

\scriptsize
\setlength{\tabcolsep}{2.0pt}
\renewcommand{\arraystretch}{1.05}

\resizebox{\textwidth}{!}{%
\begin{tabular}{@{}lccccccccccc@{}}
\toprule
\textbf{Model}
& \textbf{ArXiv}
& \textbf{Doc}
& \textbf{Info}
& \textbf{TabFQuAD}
& \textbf{TAT}
& \textbf{ShiftProj}
& \textbf{Syn-AI}
& \textbf{Syn-Energy}
& \textbf{Syn-Gov}
& \textbf{Syn-Health}
& \textbf{Avg.} \\
\midrule

\multicolumn{12}{c}{\textit{Multi-Vector Retrieval}} \\
\midrule
ColPali-v1.3
& $83.3$ & $58.4$ & $85.5$ & $86.7$ & $70.8$ & $77.3$ & $97.4$ & $94.6$ & $96.1$ & $97.4$ & $84.8$ \\

ColQwen2.5-v0.2
& $\textbf{88.9}$ & $\textbf{63.6}$ & $\underline{92.5}$ & $90.8$ & $\underline{82.1}$ & $87.9$ & $\textbf{99.6}$ & $\underline{96.1}$ & $95.8$ & $98.0$ & $\underline{89.5}$ \\

ColQwen2-v1.0
& $88.0$ & $\underline{61.5}$ & $\underline{92.5}$ & $89.0$ & $\textbf{82.2}$ & $\underline{89.9}$ & $99.0$ & $95.9$ & $95.5$ & $\underline{98.8}$ & $89.2$ \\

Jina(LI)
& $\underline{88.5}$ & $60.1$ & $\textbf{93.8}$ & $\textbf{95.6}$ & $80.3$ & $\textbf{92.9}$ & $\underline{99.3}$ & $\textbf{97.3}$ & $\underline{96.6}$ & $\textbf{99.1}$ & $\textbf{90.4}$ \\

\midrule
\multicolumn{12}{c}{\textit{Single-Vector Baselines}} \\
\midrule
SigLIP
& $50.2$ & $31.3$ & $69.7$ & $60.3$ & $27.5$ & $25.0$ & $67.8$ & $73.5$ & $75.3$ & $83.1$ & $56.4$ \\

VLM2Vec
& $42.8$ & $26.7$ & $66.7$ & $63.5$ & $21.4$ & $25.0$ & $53.5$ & $63.5$ & $64.0$ & $70.7$ & $49.8$ \\

Jina
& $84.8$ & $51.8$ & $87.0$ & $95.1$ & $65.6$ & $82.0$ & $96.8$ & $90.0$ & $93.8$ & $96.5$ & $84.3$ \\

Eager
& $85.3$ & $46.1$ & $87.2$ & $89.6$ & $62.0$ & $73.1$ & $97.5$ & $94.3$ & $93.1$ & $97.0$ & $82.5$ \\

MoCa
& $86.5$ & $55.8$ & $\svbest{89.5}$ & $93.3$ & $65.8$ & $82.6$ & $97.7$ & $94.0$ & $96.0$ & $95.6$ & $\svbest{85.7}$ \\

\midrule
\multicolumn{12}{c}{\textit{Ours: MINER Single-Vector Retrieval}} \\
\midrule
\textbf{MINER-Jina}
& $84.6\nogain{-0.2\%}$
& $52.6\gain{1.5\%}$
& $86.7\nogain{-0.3\%}$
& $\svbest{\underline{95.2}\gain{0.1\%}}$
& $64.3\nogain{-2.0\%}$
& $82.1\gain{0.1\%}$
& $97.1\gain{0.3\%}$
& $90.5\gain{0.6\%}$
& $93.9\gain{0.1\%}$
& $96.8\gain{0.3\%}$
& $84.4\gain{0.1\%}$ \\

\textbf{MINER-Eager}
& $\svbest{86.8\gain{1.8\%}^*}$
& $\svbest{57.8\gain{25.4\%}^*}$
& $88.2\gain{1.1\%}^*$
& $90.4\gain{0.9\%}^*$
& $63.3\gain{2.1\%}$
& $77.1\gain{5.5\%}^*$
& $97.7\gain{0.2\%}$
& $\svbest{95.2\gain{1.0\%}}$
& $94.3\gain{1.3\%}$
& $\svbest{97.1\gain{0.1\%}}$
& $84.8\gain{2.8\%}^*$ \\

\textbf{MINER-MoCa}
& $86.3\nogain{-0.2\%}$
& $55.3\nogain{-0.9\%}$
& $88.8\nogain{-0.8\%}$
& $93.0\nogain{-0.3\%}$
& $\svbest{66.9\gain{1.7\%}^*}$
& $\svbest{82.7\gain{0.1\%}}$
& $\svbest{98.5\gain{0.8\%}}$
& $93.3\nogain{-0.7\%}$
& $\svbest{\textbf{96.7}\gain{0.7\%}}$
& $94.8\nogain{-0.8\%}$
& $85.6\nogain{-0.1\%}$ \\

\bottomrule
\end{tabular}
}
\end{table*}

\begin{table*}[!h]
\centering
\caption{
NDCG@$10$ on ViDoRe V$3$.
\textbf{Bold} indicates the best score in each column and \underline{underline} indicates the second-best score.
\colorbox{svbestblue}{Blue shading} denotes the best single-vector retriever in each column.
$^*$ indicates that MINER has statistically significant improvement over its corresponding backbone.
}
\label{tab:vidore-v3-main}

\scriptsize
\setlength{\tabcolsep}{2.0pt}
\renewcommand{\arraystretch}{1.05}

\resizebox{\textwidth}{!}{%
\begin{tabular}{@{}lccccccccc@{}}
\toprule
\textbf{Model}
& \textbf{CS}
& \textbf{Energy}
& \textbf{Fin-EN}
& \textbf{Fin-FR}
& \textbf{HR}
& \textbf{Industrial}
& \textbf{Pharma}
& \textbf{Physics}
& \textbf{Avg.} \\
\midrule

\multicolumn{10}{c}{\textit{Multi-Vector Retrieval}} \\
\midrule
ColPali
& $65.3$ & $46.9$ & $34.4$ & $21.8$ & $44.8$ & $35.6$ & $53.1$ & $41.7$ & $43.0$ \\

ColQwen2
& $68.6$ & $48.6$ & $39.0$ & $20.0$ & $45.1$ & $38.3$ & $52.2$ & $41.6$ & $44.2$ \\

ColQwen2.5
& $\textbf{72.3}$ & $\underline{59.5}$ & $\underline{52.3}$ & $\underline{39.1}$ & $51.2$ & $\underline{41.3}$ & $57.9$ & $45.9$ & $\underline{52.4}$ \\

Jina(LI)
& $\underline{71.8}$ & $\textbf{63.5}$ & $\textbf{59.3}$ & $\textbf{46.1}$ & $\textbf{59.5}$ & $\textbf{50.4}$ & $\textbf{63.1}$ & $46.6$ & $\textbf{57.5}$ \\

\midrule
\multicolumn{10}{c}{\textit{Dense / Single-Vector Baselines}} \\
\midrule
Jina
& $67.4$ & $55.3$ & $44.5$ & $31.9$ & $51.9$ & $39.0$ & $58.0$ & $44.2$ & $49.0$ \\

Eager
& $69.6$ & $53.9$ & $42.1$ & $28.9$ & $48.5$ & $33.6$ & $59.5$ & $44.8$ & $47.6$ \\

MoCa
& $68.6$ & $54.1$ & $\svbest{49.3}$ & $\svbest{33.1}$ & $47.3$ & $38.5$ & $54.6$ & $\underline{46.9}$ & $49.1$ \\

\midrule
\multicolumn{10}{c}{\textit{Ours: MINER Single-Vector Retrieval}} \\
\midrule
\textbf{MINER-Jina}
& $68.4\gain{1.5\%}^*$
& $\svbest{56.4\gain{2.0\%}^*}$
& $44.9\gain{0.9\%}^*$
& $32.4\gain{1.6\%}^*$
& $\svbest{\underline{52.1}\gain{0.4\%}}$
& $\svbest{39.8\gain{2.1\%}^*}$
& $58.9\gain{1.6\%}^*$
& $44.9\gain{1.6\%}^*$
& $\svbest{49.7\gain{1.4\%}^*}$ \\

\textbf{MINER-Eager}
& $\svbest{70.5\gain{1.3\%}^*}$
& $55.5\gain{3.0\%}^*$
& $43.8\gain{4.0\%}^*$
& $29.7\gain{2.8\%}^*$
& $49.6\gain{2.3\%}^*$
& $35.7\gain{6.3\%}^*$
& $\svbest{\underline{60.1}\gain{1.0\%}^*}$
& $45.4\gain{1.3\%}^*$
& $48.8\gain{2.5\%}^*$ \\

\textbf{MINER-MoCa}
& $70.1\gain{2.2\%}^*$
& $54.9\gain{1.5\%}^*$
& $49.1\nogain{-0.4\%}$
& $32.9\nogain{-0.6\%}$
& $47.7\gain{0.8\%}$
& $38.8\gain{0.8\%}$
& $55.4\gain{1.5\%}^*$
& $\svbest{\textbf{47.3}\gain{0.9\%}^*}$
& $49.5\gain{0.8\%}^*$ \\

\bottomrule
\end{tabular}
}
\end{table*}

\FloatBarrier
\newpage
\section{Full Efficiency Analysis}

\label{app:Efficiency}

\begin{table}[h!]
\centering
\caption{
Per-dataset efficiency and retrieval-quality results on ViDoRe v2 using Jina.
QPS and storage are measured with Qdrant~\citep{qdrant} for each subset. NDCG@5 follows the main ViDoRe v2 evaluation.
Higher QPS and NDCG@5 are better; lower storage is better.
\textsc{MINER} remains single-vector retrieval with the same final embedding size and storage footprint as the dense Jina baseline.
} 
\label{tab:appendix-efficiency-v2}

\small
\setlength{\tabcolsep}{4.5pt}
\renewcommand{\arraystretch}{1.12}

\resizebox{0.7\textwidth}{!}{%
\begin{tabular}{@{}llccc@{}}
\toprule
\textbf{Dataset}
& \textbf{Method}
& \textbf{QPS} $\uparrow$
& \textbf{Storage (MB)} $\downarrow$
& \textbf{NDCG@5} $\uparrow$ \\
\midrule

\multirow{3}{*}{Bio}
& Jina & $15.21$ & $19.89$ & $58.5$ \\
& Jina(LI) & $2.46$ & $859.64$ & $60.9$ \\
& \textsc{MINER}-Jina & $14.31$ & $19.89$ & $58.6$ \\

\midrule

\multirow{3}{*}{Econ}
& Jina & $16.26$ & $8.85$ & $55.1$ \\
& Jina(LI) & $4.80$ & $378.74$ & $51.9$ \\
& \textsc{MINER}-Jina & $14.97$ & $8.85$ & $56.8$ \\

\midrule

\multirow{3}{*}{ESG-Human}
& Jina & $15.16$ & $30.11$ & $54.8$ \\
& Jina(LI) & $1.99$ & $1267.57$ & $65.1$ \\
& \textsc{MINER}-Jina & $14.12$ & $30.11$ & $56.8$ \\

\midrule

\multirow{3}{*}{ESG-Syn}
& Jina & $15.01$ & $30.11$ & $44.8$ \\
& Jina(LI) & $1.75$ & $1267.57$ & $52.5$ \\
& \textsc{MINER}-Jina & $14.06$ & $30.11$ & $48.7$ \\

\midrule

\multirow{3}{*}{Avg.}
& Jina & $15.41$ & $22.24$ & $53.3$ \\
& Jina(LI) & $2.75$ & $943.38$ & $57.6$ \\
& \textsc{MINER}-Jina & $14.36$ & $22.24$ & $55.2$ \\

\bottomrule
\end{tabular}%
}
\end{table}

Table~\ref{tab:appendix-efficiency-v2} provides the per-dataset breakdown behind the aggregate efficiency results in Section~\ref{sec:efficiency}. 
Across all ViDoRe V$2$ subsets, \textsc{MINER} has the same storage footprint as the dense Jina baseline because it preserves single-vector retrieval with the same final embedding dimensionality. 
Its QPS is consistently slightly lower than dense retrieval, reflecting the modest overhead of computing the fused representation, but the gap remains small and stable across datasets. 
In contrast, late interaction incurs substantially larger and more variable storage and latency costs because it indexes multiple visual token embeddings per document, whose number can vary with document content and rendered page resolution. 
This leads to much lower QPS and more than $40\times$ larger index storage on average.

The retrieval-quality gains of \textsc{MINER} are also consistent across datasets. 
Compared with the dense Jina baseline, \textsc{MINER} improves nDCG@$5$ on every ViDoRe V$2$ subset, with gains of $0.1$, $1.7$, $2.0$, and $3.9$ on Bio, Econ, ESG-Human, and ESG-Syn, respectively. 
These results show that the aggregate improvement in Figure~\ref{fig:efficiency} is not driven by a single dataset, but instead reflects a stable quality-efficiency trade-off across the benchmark.

\section{Neuron Selection Analysis}
\label{app:Neuron_select_plots}

\begin{figure}[H]
  \centering
  \resizebox{0.75\linewidth}{!}{%
    \includegraphics{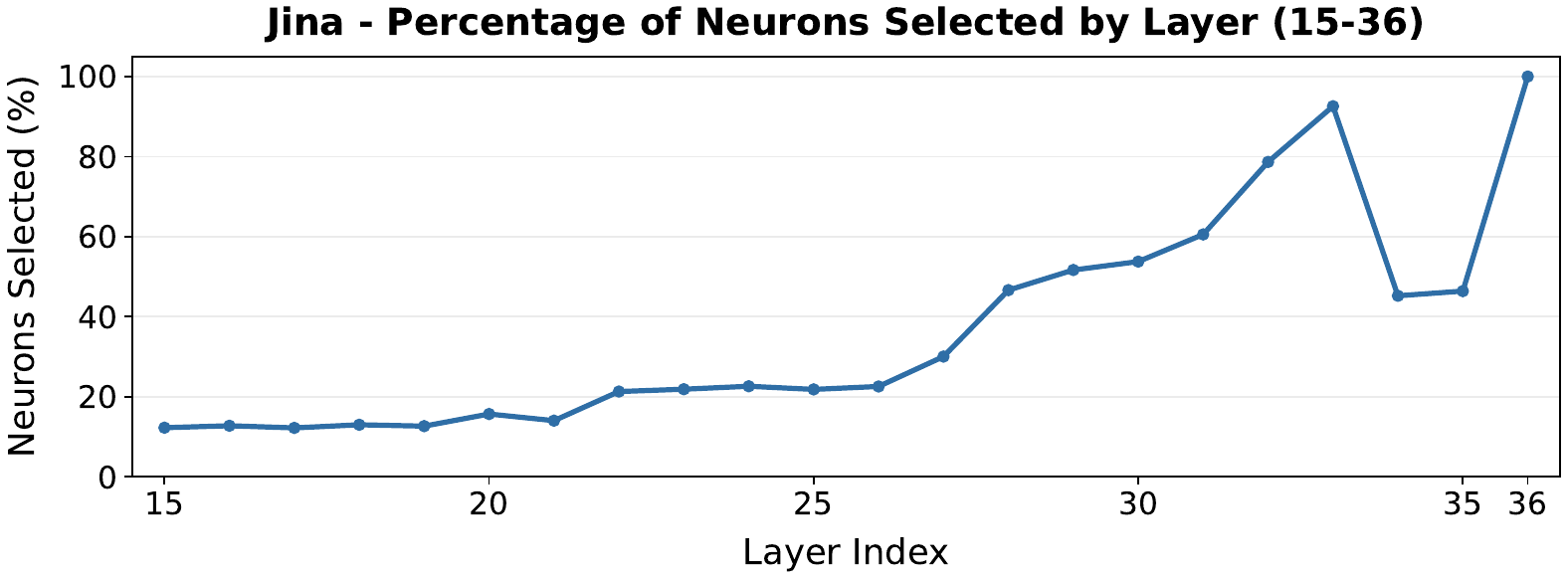}%
  }
  \caption{
  Percentage of neurons selected by the Top-$p$ masking procedure for Jina across layers 15--36.
  }
  \label{fig:jina_neuron_selection}
\end{figure}

\begin{figure}[H]
  \centering
  \resizebox{0.75\linewidth}{!}{%
    \includegraphics{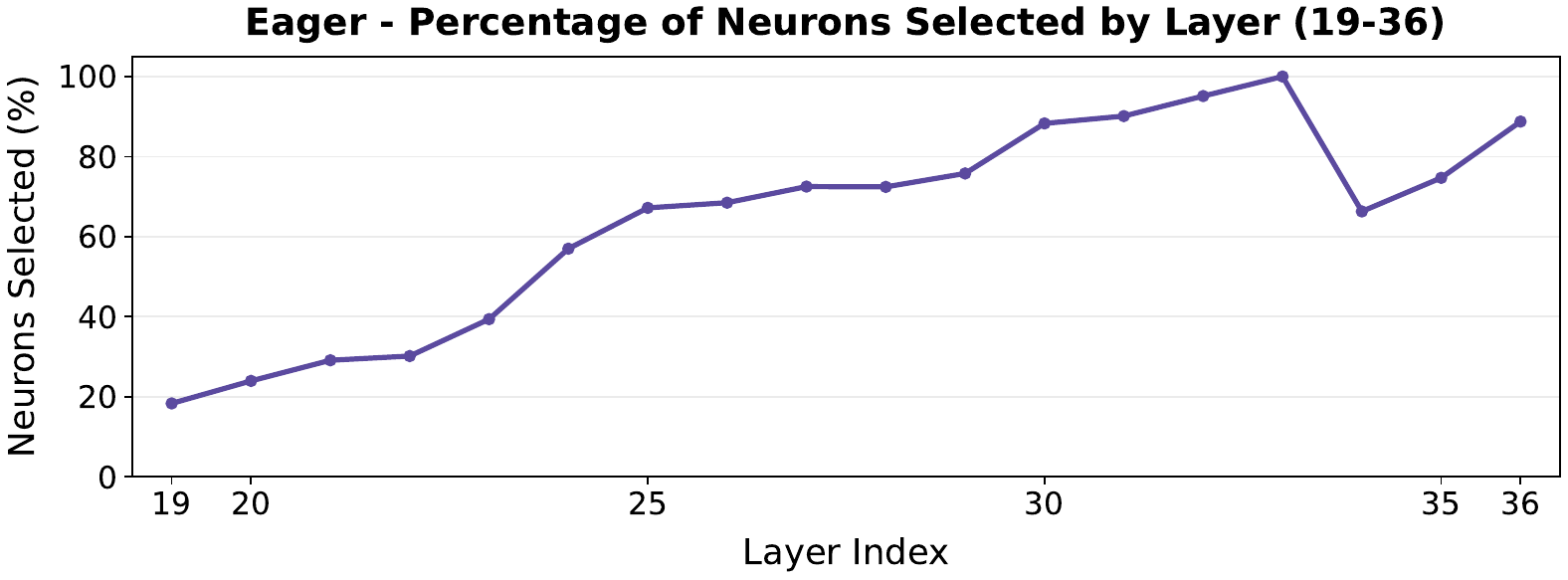}%
  }
  \caption{
  Percentage of neurons selected by the Top-$p$ masking procedure for Eager across layers 19--36.
  }
  \label{fig:eager_neuron_selection}
\end{figure}

\begin{figure}[H]
  \centering
  \resizebox{0.75\linewidth}{!}{%
    \includegraphics{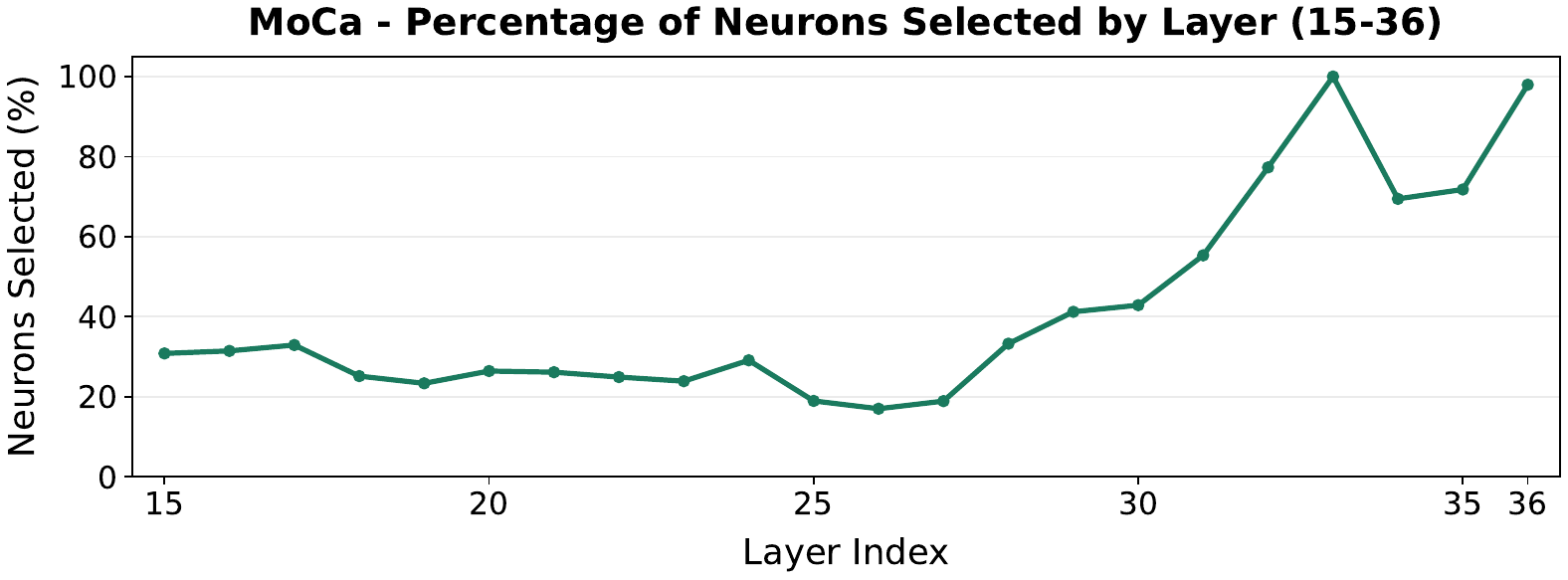}%
  }
  \caption{
  Percentage of neurons selected by the Top-$p$ masking procedure for MoCa across layers 15--36.
  }
  \label{fig:moca_neuron_selection}
\end{figure}

Figure~\ref{fig:jina_neuron_selection}, Figure~\ref{fig:eager_neuron_selection}, and Figure~\ref{fig:moca_neuron_selection} show the percentage of neurons retained by the Top-$p$ masking procedure at each selected internal layer under the hyperparameter $\rho = 0.2$ and $\tau_{\mathrm{CKA}} = 0.6$. 
Across all three backbones, the neuron-selection profiles are highly similar, despite differences in model training, backbone design, and readout mechanism. 
In particular, the retained fraction is relatively low in earlier and middle layers, increases sharply in the late layers, and exhibits a consistent drop around layer 34 before rising again near the final layer. 
This shared structure suggests that the masking procedure is not simply fitting backbone-specific noise, but instead captures a recurring pattern in how useful cross-modal information is distributed across internal representations. 
Earlier and middle layers appear to contain useful but sparse features, requiring selective extraction, whereas the final layers are more directly aligned with the dense embedding interface and therefore retain substantially more dimensions. 
The consistent dip around layer 34 further indicates that high layer index alone is not sufficient: even among late layers, some representations contain more redundant information than neighboring layers.


\newpage

\section{Additional Sensitivity Analysis}

\label{app:sensitivity}
\begin{figure}[H]
    
    \centering
    \includegraphics[width=\linewidth]{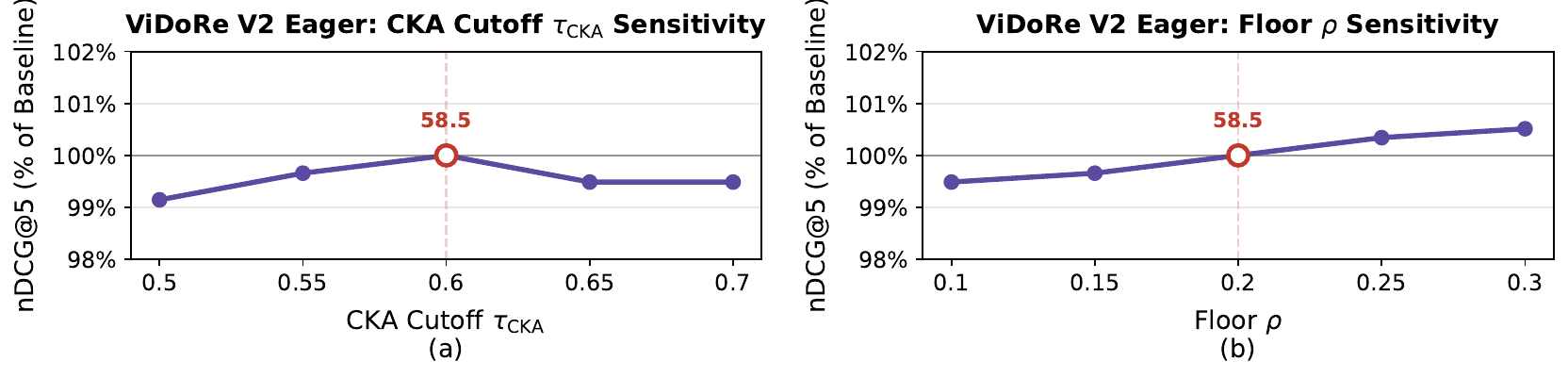}
    \caption{
    Hyperparameters sensitivity analysis of Eager, reported as a percentage of the default configuration
    }
    \label{fig:eager_sensitivity_checks}
\end{figure}

\begin{figure}[H]
    \centering
    
    \includegraphics[width=\linewidth]{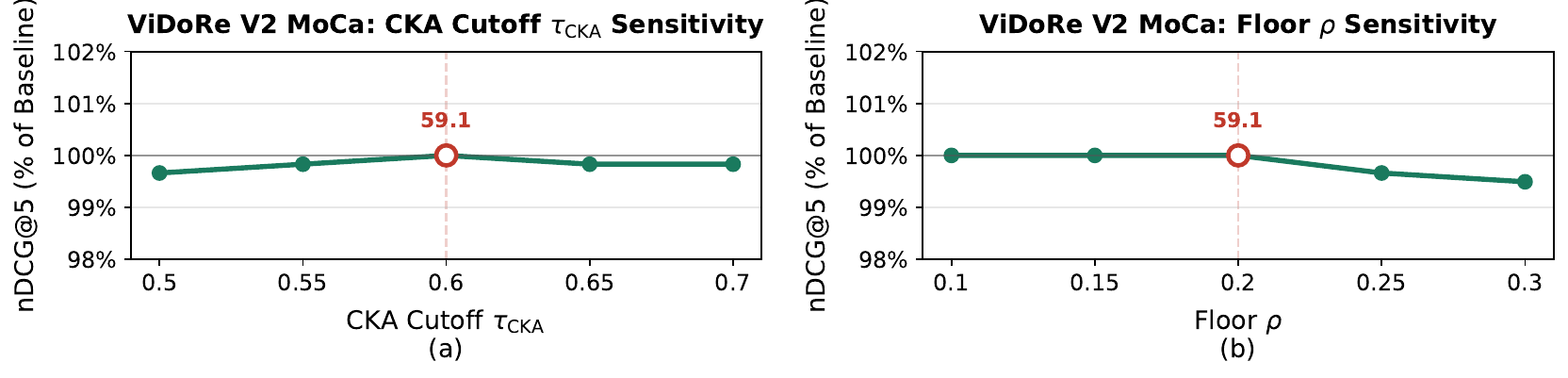}
    \caption{
    Hyperparameters sensitivity analysis of MoCa, reported as a percentage of the default configuration
    }
    \label{fig:moca_sensitivity_checks}
\end{figure}

Table~\ref{tab:cka_cutoff_scand} reports the candidate layer ranges selected under different CKA cutoff values for each backbone. 
As the cutoff increases, the selected range becomes progressively more restrictive and shifts toward later layers, reflecting the stronger alignment of late-layer representations with the final embedding space. 
Despite this change in the number of candidate layers, the sensitivity plots in Figures~\ref{fig:eager_sensitivity_checks} and~\ref{fig:moca_sensitivity_checks} show that retrieval performance remains stable across the tested cutoffs and sparsity floors. 
This indicates that \textsc{MINER} does not rely on a fragile layer-selection cutoff, but instead remains robust across a reasonable range of CKA-based candidate sets.

\begin{table}[!h]
\centering
\caption{Candidate layer ranges under different CKA cutoffs.}
\label{tab:cka_cutoff_scand}
\small
\setlength{\tabcolsep}{8pt}
\renewcommand{\arraystretch}{0.95}
\begin{tabular}{lcc}
\toprule
\textbf{Backbone} & \textbf{CKA cutoff} & $\boldsymbol{S_{\mathrm{cand}}}$ \\
\midrule
\multirow{5}{*}{Jina}
& $0.50$ & $12$--$36$ \\
& $0.55$ & $13$--$36$ \\
& $0.60$ & $15$--$36$ \\
& $0.65$ & $22$--$36$ \\
& $0.70$ & $28$--$36$ \\
\midrule
\multirow{5}{*}{Eager}
& $0.50$ & $10$--$36$ \\
& $0.55$ & $12$--$36$ \\
& $0.60$ & $19$--$36$ \\
& $0.65$ & $21$--$36$ \\
& $0.70$ & $22$--$36$ \\
\midrule
\multirow{5}{*}{MoCa}
& $0.50$ & $12$--$36$ \\
& $0.55$ & $13$--$36$ \\
& $0.60$ & $15$--$36$ \\
& $0.65$ & $19$--$36$ \\
& $0.70$ & $21$--$36$ \\
\bottomrule
\end{tabular}
\end{table}

\section{Limitation} \label{app:limitation}
A limitation of our current evaluation is that MINER requires access to internal hidden states, and our controlled training protocol requires access to the training data associated with each backbone. This is intentional: to ensure that improvements come from better utilization of internal representations rather than from introducing additional supervision, we train the MINER module only on the original training data of the corresponding backbone. As a result, we do not evaluate on closed-source models where hidden states are inaccessible, or on strong open-weight models whose training data are not fully disclosed, such as Qwen/Qwen3-VL-Embedding-8B~\citep{li2026qwen3vlembedding}. Nevertheless, this is an evaluation constraint rather than a methodological restriction: when hidden states and appropriate training data are available, MINER can be applied as a lightweight plug-in module without modifying the backbone architecture or changing the final dense retrieval interface.


\end{document}